%% file: main.tex
\documentclass{article}

\usepackage[preprint]{neurips_2024}

\usepackage[utf8]{inputenc} 
\usepackage[T1]{fontenc}    
\usepackage{hyperref}       
\usepackage{url}            
\usepackage{booktabs}       
\usepackage{amsfonts}       
\usepackage{nicefrac}       
\usepackage{microtype}      
\usepackage{xcolor}         

\usepackage{amsmath}
\usepackage{graphicx}
\usepackage{amsthm}
\usepackage{amssymb}
\usepackage{multirow}
\usepackage{wrapfig,lipsum}
\usepackage{algorithm}
\usepackage{enumitem}
\usepackage{algpseudocode}
\usepackage{mathrsfs}
\usepackage{ulem}
\usepackage{comment}
\usepackage{tabularx}
\usepackage{booktabs}

\hypersetup{
    colorlinks=true,
    linkcolor=red,
    filecolor=black,
    citecolor=blue,
    urlcolor=red,
}
\newcommand{\blfootnote}[1]{\begingroup
\renewcommand\thefootnote{}\footnote{#1}\addtocounter{footnote}{-1}
\endgroup}
\newcommand{\method}{\textit{UKnow }}
\usepackage{colortbl}
\newlength\savewidth\newcommand\shline{\noalign{\global\savewidth\arrayrulewidth
  \global\arrayrulewidth 1pt}\hline\noalign{\global\arrayrulewidth\savewidth}}

\newcommand{\tablestyle}[2]{\setlength{\tabcolsep}{#1}\renewcommand{\arraystretch}{#2}\centering\footnotesize}
\usepackage{bbding}

\definecolor{baseColor}{rgb}{0.75,0.05,0.1}
\newcommand{\base}[1]{{\color{baseColor}#1}}
\definecolor{checkmarkColor}{rgb}{0.1,0.75,0.1}
\newcommand{\checkc}[1]{{\color{checkmarkColor}#1}}
\definecolor{demphcolor}{RGB}{144,144,144}

\definecolor{mygray}{gray}{0.4}
\usepackage{pifont}
\usepackage{subfigure} 
\newcommand\hl{\bgroup\markoverwith
{\textcolor[RGB]{239,239,239}{\rule[-.5ex]{2pt}{2.5ex}}}\ULon}
\usepackage{amsfonts}
\usepackage{bm}
\usepackage{bbm}
\usepackage{tikz}
\usepackage{booktabs} 
\usepackage{lipsum}
\usepackage{listings}

\title{UKnow: A Unified Knowledge Protocol with Multimodal Knowledge Graph Datasets for Reasoning and Vision-Language Pre-Training}

\author{Biao Gong$^{1}$, 
Shuai Tan$^{2}$,
Yutong Feng$^{1}$,
Xiaoying Xie$^{1}$,\\
\textbf{
Yuyuan Li$^{3,4\dagger}$, 
Chaochao Chen$^{4}$,
Kecheng Zheng$^{2}$,
Yujun Shen$^{2}$,
Deli Zhao$^{1}$,
}\\
$^1$Alibaba Group, $^2$Ant Group, $^3$Hangzhou Dianzi University, $^4$Zhejiang University \\
\\[-1.5ex]
\small{\texttt{\{a.biao.gong, tanshuai2001, fengyutong.fyt\}@gmail.com\ souyu.xxy@alibaba-inc.com}}\\[-0.6ex]
\small{\texttt{y2li@hdu.edu.cn\ zjuccc@zju.edu.cn\ \{zkechengzk, shenyujun0302, zhaodeli\}@gmail.com}}\\
}

\begin{document}
\blfootnote{$\dagger$ Corresponding Author.}

\maketitle

\vspace{-5mm}
\begin{abstract}
This work presents a unified knowledge protocol, called \textbf{\textit{UKnow}}, which facilitates knowledge-based studies from the perspective of data.
Particularly focusing on visual and linguistic modalities, we categorize data knowledge into five unit types, namely, in-image, in-text, cross-image, cross-text, and image-text, and set up an efficient pipeline to help construct the multimodal knowledge graph from any data collection.
Thanks to the logical information naturally contained in knowledge graph, organizing datasets under \method format opens up more possibilities of data usage compared to the commonly used image-text pairs.
Following \method protocol, we collect, from public international news, a large-scale multimodal knowledge graph dataset that consists of 1,388,568 nodes (with 571,791 vision-related ones) and 3,673,817 triplets.
The dataset is also annotated with rich event tags, including 11 coarse labels and 9,185 fine labels.
Experiments on 4 benchmarks demonstrate the potential of \method in supporting common-sense reasoning and boosting vision-language pre-training with a single dataset, benefiting from its unified form of knowledge organization. See Appendix~\ref{sec: addotional_data} to download the dataset.
\end{abstract}

\input{body/1_intro}

\input{body/2_rela}

\input{body/4_method}

\input{body/5_exp}

\input{body/6_con}


\small

\input{main.bbl}
\bibliographystyle{plain}
\bibliography{main}

\newpage
\appendix
\input{body/7_app}

\end{document}

%% file: body/1_intro.tex
\section{Introduction}
\label{sec:intro}

Recent efforts have been attracted to leverage the \textit{multimodal knowledge graph}~\cite{multi} for data-driven intelligence.
Inspired by the human mastery knowledge network~\cite{knowledgesur}, we consider that the multimodal knowledge graph, which naturally accommodates heterogeneous data based on its format of complex network~\cite{mmkgr,wang2019multimodal}, is well suited for constructing a unified knowledge criterion from the perspective of data.
Driven by the multimodal knowledge graph, models can easily introduce external knowledge~\cite{knowledge}, discover long-range relations~\cite{NELL995} and understand more logical semantics~\cite{yago}.
However, existing datasets of the multimodal knowledge graph commonly focus on only one task like common-sense reasoning~\cite{wn9,FBTXTIMG} due to their limited scale and irregular data organization. Therefore, it is imperative to construct a well-organized multimodal knowledge graph dataset with large-scale and rich-logic, which enables delving into deeper foundational problems in lower layers, such as the knowledge based vision-language pre-training.

\begin{wrapfigure}{r}{0.5\textwidth} %
\vspace{-1mm}
\includegraphics[width=1.0\linewidth]{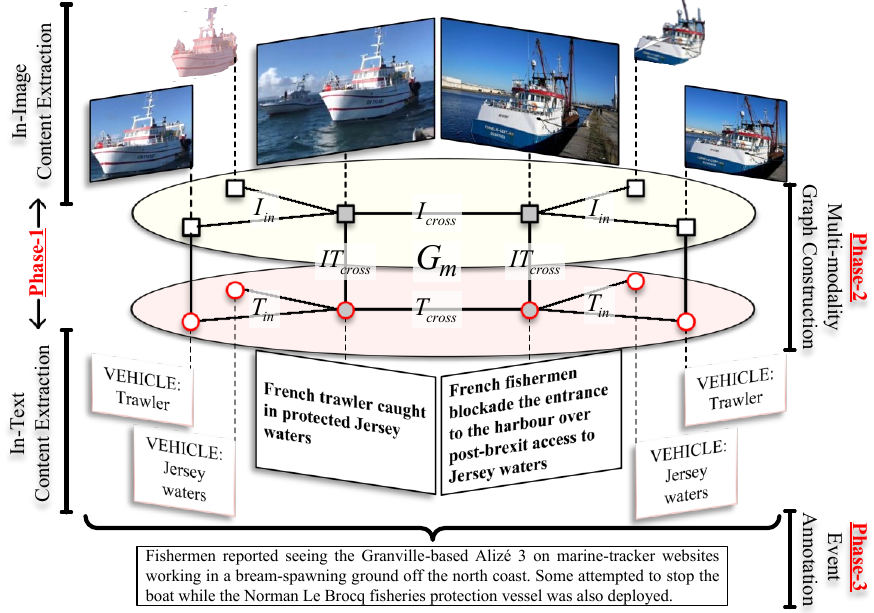}
\vspace{-3mm}
     \caption{
        \textbf{Overview of \method protocol}, consisting of five unit knowledge types, namely, 
in-image~$I_{in}$ (\textit{e.g.}, object), in-text~$T_{in}$ (\textit{e.g.}, entity), cross-image~$I_{cross}$ (\textit{e.g.}, image similarity), cross-text~$T_{cross}$ (\textit{e.g.}, text continuity), and image-text~$IT_{cross}$ (\textit{e.g.}, description). 
    }
	\label{fig:1}
\vspace{-0.2in}
\end{wrapfigure}

To this end, we propose \textbf{\textit{UKnow}}, a \textbf{U}nified \textbf{Know}ledge protocol, which facilitates knowledge-based studies~\cite{BetaE_KGreasoning,kbVQA,cross} from the perspective of data.
Particularly focusing on visual and linguistic modalities, we categorize data knowledge into five unit types, namely, 
in-image~$I_{in}$, in-text~$T_{in}$, cross-image~$I_{cross}$, cross-text~$T_{cross}$, and image-text~$IT_{cross}$. 
As shown in Fig.~\ref{fig:1}, these knowledge types are together named as \textit{Knowledge-View} which can be easily used to construct a multimodal knowledge graph ($\mathbf{G}_m$).

To verify that \method can serve as a standard protocol, we further set up an efficient data processing pipeline, consisting of \textit{Phase-1/2/3}, to reorganize existing datasets into \textit{UKnow}'s format. Please note that, this pipeline is also able to automatically extend an existing image-text dataset like LAION-5B~\cite{schuhmann2022laion} with more useful information to build a new dataset. A brief description of each \textit{Phase} is as follows:

\textbf{\textit{Phase-1: }}\textit{Content Extraction}. We use pre-trained models to preprocess data and extract useful content. Note that pre-trained models can be replaced / added / disabled freely as needed. \textbf{\textit{Phase-2: }}\textit{Information Symbolization}. Since the results obtained in \textit{Phase-1} (\textit{e.g.}, images and texts) cannot be used directly for graph construction, we adopt information symbolization strategy to arrange all of them into the index in this phase. This information symbolization strategy numbers all original or generated data by a certain rule, which links the nodes from \textit{Phase-1} to make a multimodal graph. \textbf{\textit{Phase-3: }}\textit{Knowledge Construction}. Two kinds of internal knowledge ($I_{in}, T_{in}$) and three kinds of associative knowledge ($I_{cross}, T_{cross}, IT_{cross}$) are aggregated into one graph ($\mathbf{G}_m$) in this phase as shown in Fig.~\ref{fig:1}.

 Following \method protocol and above pipeline, we build a novel large-scale multimodal knowledge graph. Considering that a large-scale event dataset is of practical significance for real-world applications, such as information retrieval and public sentiment analysis, our data are collected from public international news. Overall, our dataset contains 1,388,568 nodes of which 571,791 are vision relevant (\textit{i.e.}, news images or visual objects). The number of triples in the entire graph is 3,673,817. To the best of our knowledge, this dataset has become the largest multimodal knowledge graph dataset of international news events. 
Moreover, to organize data in a more structured way and enhance dataset with more category labels, our dataset introduces a \textit{hierarchical event annotation} for each news, including \textbf{\textit{Event-11}} and \textbf{\textit{Event-9185}}. Specifically, the former contains general event categories such as \textit{``Sports, Ceremony, ...''}, while the latter consists of real human activity in the history such as \textit{``2019 NBA All-Star Game, 2019 Daytona 500, ...''}.
More details about the annotation are shown in Sec.~\ref{sec:da}, Fig~\ref{fig:5}, and Tab.~\ref{tab:coraevent}.

In summary, our \textbf{contributions} are as follows:

\begin{itemize}[labelsep=0.4em, leftmargin=1em,itemindent=0em]
    \vspace{-8pt}
    \setlength{\itemsep}{0pt}
    \setlength{\parsep}{0pt}
    \setlength{\parskip}{0pt}
\item We propose \method to introduce the multimodal knowledge graph into the vision field as a new standard of data organization, which features the relation inside data in addition to the original data format. Such a protocol opens up the possibilities of data usage such that more logic-rich downstream tasks can be expected in the future.  %
\item We design an efficient data processing pipeline for constructing dataset following our \method protocol, together with a large-scale multimodal knowledge graph dataset collected from public international news. We also equip the dataset with hierarchical event annotations, which can help models understand human activities and history. See Appendix~\ref{sec: addotional_data} to download the dataset.
\item We provide some examples of the usage of \method in practical applications. Experiments on four benchmarks showcase the advantages of \method in supporting common-sense reasoning and boosting vision-language pre-training with a unified form of data organization, making it possible to evaluate various tasks on a single dataset.
\end{itemize}

%% file: body/2_rela.tex
\section{Related Work}\label{sec:related-work}
\subsection{Existing Knowledge Representation Formats}\label{sec:triples}
In recent years, a growing abundance multi-modal data are disseminated, linking diverse information across various modalities such as text and image in a global data space. This interconnected web of heterogeneous data constitutes a vast repository of information termed as knowledge. With the development of large-scale models, the utilization of knowledge has seen a notable surge in exploration.
Existing knowledge-based deep learning models are broadly divided into two aspects: (1) external knowledge introduction~\cite{kldrivenbenchmarking}, (2) internal knowledge mining~\cite{jing2020self}. The former leverages expert knowledge by introducing external data~\cite{krisp,lauscher2020common,chen2020recall} or pre-trained models~\cite{cris,ruta2022stylebabel,esmaeilpour2022zero,yang2022empirical}. The latter means constructing correlations of training data by similarity~\cite{pan2020self,guo2022contrastive,han2020self} or discovering favorable substructures of internal models~\cite{glip,chi2021improving,clipEvent,wei2021aligning}.

However, from the perspective of data organization, existing studies often claim to be knowledge-based only using one piece of them, which is actually incomplete and cannot be analogous to the complex knowledge network held by humans. In this work, we build a unified knowledge protocol based on the multimodal knowledge graph to define the unified knowledge on multimodal data.

\subsection{Multimodel Knowledge Graph Datasets}
\label{sec:triples2}

The Multimodal Knowledge Graph (MMKG) serves as a potent means to store and leverage multimodal knowledge explicitly, which bolsters and enhances model performances across diverse domains. In Tab.~\ref{tab:tab1}, we list mainstream multimodal knowledge graph datasets~\cite{richpedia,imageGraph,visualGenome,visualSem,gaia,wn9,mmkg,resin,xu2022relation, li2023vision, chen2023rethinking, zhang2023aspectmmkg, wang2023tiva, zha2023m2conceptbase, lee2023vista}, constructed by texts and images with detailed information.
In terms of data scale, VisualGenome~\cite{visualGenome} is a multimodal knowledge graph which contains 40,480 relations, 108,077 image nodes with objects. The ImageGraph~\cite{imageGraph} further pushed up the number of image nodes to 829,931 but missing the extraction of visual objects. Recently, VisualSem~\cite{visualSem} implements a multimodal knowledge graph with 938$K$ image nodes and 89,896 entity nodes, but it only uses 15 types of relation to build the graph. On the route of increasing the number of entity nodes, while Multi-OpenEA~\cite{li2023vision} boasts 920,000 entity nodes, surpassing prior methods, our endeavor has achieved 1,388,568 nodes, establishing the largest graph thus far.
Besides, most of existing multimodal knowledge graphs are more like a vision-similarity-based image library~\cite{liu2016deepfashion,song2021matching} with image descriptions and meta information, it lacks the most valuable feature of the knowledge graph: ``The Logical Connection''. This logic refers to the additional association between two nodes that were originally unrelated, triggered by a news event involving these two nodes. For example, prior to the news event "Celebrity 1 visits Area 1," there was no relation between Celebrity 1 and the Area 1. The newly added "visit" relation in
{\footnotesize $<$(``$\mathtt{Celebrity 1}$''), $\mathtt{visit}$, (``$\mathtt{Area 1}$'')$>$} tuple exemplifies this logic, which is highly beneficial for downstream tasks.

\begin{table}[t]
     \caption{\textbf{Statistics} of various multimodal knowledge graph datasets. \textbf{TRIPLE} is the basic component of knowledge graph (Sec.~\ref{sec:triples}), \textbf{WEB} and \textbf{GIT} indicate homepage and Github repository respectively. \textbf{EVENT} indicates the news event.}  
      \vspace{2mm}

     \resizebox{\linewidth}{!}{
     
      \begin{tabular}{cccc|ccc|cc|c}
        \bf DATASET &\bf YEAR& \bf MULTIMODAL INFO.  & \bf SOURCE  & \bf NODE & \bf IMAGE & \bf TRIPLE & \bf WEB & \bf GIT& \bf EVENT \\
    \shline
    WN9-IMG-TXT~\cite{wn9} & 2016 & ENT. &WN18, ImageNet& 6,555 & 63,225 & 14,397 &  & $\checkmark$&\base{\tiny \XSolidBrush}\\
    ImageGraph~\cite{imageGraph} & 2017 & ENT./CONCEPT &FB15k& 14,870 & 829,931 & 564,010 &  & $\checkmark$&\base{\tiny \XSolidBrush}\\
    VisualGenome~\cite{visualGenome} & 2017 & ENT. &MSCOCO& 75,729 & 108,077 & 1,531,448 & $\checkmark$ & &\base{\tiny \XSolidBrush}\\
    GAIA~\cite{gaia} & 2018 & ENT./CONCEPT &Freebase, Geonames& 457,000 & - & 38,000 &  & $\checkmark$&\base{\tiny \XSolidBrush}\\
    MMKG-FB15k~\cite{mmkg} & 2019 & ENT./CONCEPT &FB15k, Search Engine& 14,951 & 13,444 & 592,213 & $\checkmark$  & $\checkmark$&\base{\tiny \XSolidBrush}\\
    MMKG-DB15k~\cite{mmkg} & 2019 & ENT./CONCEPT &DB15k, Search Engine& 14,777 & 12,842 & 99,028 & $\checkmark$  & $\checkmark$&\base{\tiny \XSolidBrush}\\
    MMKG-YAGO15k~\cite{mmkg} & 2019 & ENT./CONCEPT &YAGO15k, Search Engine& 15,283 & 11,194 & 122,886 & $\checkmark$  & $\checkmark$&\base{\tiny \XSolidBrush} \\
    Richpedia~\cite{richpedia} & 2020 & ENT./REL./CONCEPT  &Wikipedia& 29,985 & 2,914,770 & 2,708,511 & $\checkmark$ & $\checkmark$&\base{\tiny \XSolidBrush}\\
    VisualSem~\cite{visualSem} & 2020 & ENT./CONCEPT &BabelNet& 89,896 & 930,000 & 1,500,000 &  & $\checkmark$&\base{\tiny \XSolidBrush}\\
    RESIN~\cite{resin} & 2021 & ENT./REL./CONCEPT &News& 51,422 & 6,399 & 150,220 & $\checkmark$ & $\checkmark$&\checkc{\small \checkmark}\\
    \hline
MKG-W~\cite{xu2022relation} & 2022 & ENT./REL./CONCEPT &Open EA~\cite{sun2020benchmarking}, Search Engine& 15,000 &  14,463 & - &  & &\base{\tiny \XSolidBrush}\\

MKG-Y~\cite{xu2022relation} & 2022 & ENT./REL./CONCEPT &Open EA, Search Engine&  15,000 &  14,244 & - &  & &\base{\tiny \XSolidBrush}\\

MMKB-DB15K~\cite{xu2022relation} & 2022 & ENT./REL./CONCEPT &Open EA, Search Engine&  12,842 &  12,818 &  - &  & &\base{\tiny \XSolidBrush}\\
    \hline
MarKG~\cite{zhang2022multimodal} & 2023 & ENT./CONCEPT &Wikidata, Search Engine&  11,292 &   76,424 &  34,420 &  & $\checkmark$&\base{\tiny \XSolidBrush}\\

Multi-OpenEA~\cite{li2023vision} & 2023 & ENT./CONCEPT &Open EA, Search Engine&  920,000 &   2,705,688 &  - &  & $\checkmark$&\base{\tiny \XSolidBrush}\\

UMVM~\cite{chen2023rethinking} & 2023 & ENT./CONCEPT &DBpedia, Multi-OpenEA& 238,208 &  1,073,671 & 982,626 &  & & \base{\tiny \XSolidBrush}\\

AspectMMKG~\cite{zhang2023aspectmmkg} & 2023 & ENT./CONCEPT &Wikipedia, Search Engine& 2,380 &  645,456 & - & & $\checkmark$&\base{\tiny \XSolidBrush}\\

TIVA-KG~\cite{wang2023tiva} & 2023 & ENT./REL./CONCEPT &Wikipedia, Search Engine& 443,580 &  1,695,688 & 1,382,358 &$\checkmark$ & &\base{\tiny \XSolidBrush}\\

VTKG-C~\cite{lee2023vista} & 2023 & ENT./CONCEPT &ConceptNet, WordNet & 43,267 &   461,007 & 111,491 & &$\checkmark$ &\base{\tiny \XSolidBrush}\\
    \hline
\hline
    \rowcolor[gray]{0.95} \textbf{\method} & 2024 & ENT./REL./CONCEPT &News, Wikipedia& \bf 1,388,568 &  \bf 1,073,671 & \bf 3,673,817 & $\checkmark$ & $\checkmark$ &\checkc{\small \checkmark}\\
     \end{tabular}
      }
    \vspace{-3mm}
     \label{tab:tab1}
\end{table}

Generally speaking, the above news refer to international news, which carries the most complex event logic as well as plentiful multimodal information~\cite{eann}. To completely exploit the advantages of multimodal knowledge graphs, building a dataset using event logic from international news is a natural approach. However, there is not yet a large multimodal knowledge graph of news events. RESIN~\cite{resin} is a recently published multimodal knowledge graph containing 24 types of entities, 46 types of relations and 67 types of events. The larger and fresher CLIP-Event~\cite{clipEvent} is a event rich dataset with 106,875 images and 187 types of events extracted by a text information extraction system~\cite{gaia,jointExtraction}. Actually, CLIP-Event is not a knowledge graph and its definition of ``event'' is not a news event but an action. In summary, one of goals of our work is to build a large, and realistic news-event rich, multimodal knowledge graph dataset from international news.

\subsection{Knowledge-based Downstream Tasks}
Thanks to the innovative unified knowledge proposed by our \method protocol, our dataset can readily accommodate a variety of downstream tasks. In this study, we opt for common-sense reasoning and vision-language pre-training as experimental domains to validate our dataset. Common-sense reasoning is an extremely popular task in the field of knowledge graph. Since our dataset is based on the knowledge graph, the performance validation on common-sense reasoning is indispensable. Moreover, the representations from Vision-Language Pre-training models are capable of diminishing the necessity for intricate task-specific architectures~\cite{bert}, which allows the knowledge to further flow into various downstream tasks. By incorporating these two tasks, we are able to maximize the assessment of the dataset's knowledge validity.

\noindent\textbf{Common-sense Reasoning.}
Common-sense reasoning means answering queries by logic permutations. The specific task in this work is the link prediction.
Various works~\cite{TransE,ComplEx,RotatE,ConvE,JointE,AcrE}
achieve reasoning by embedding entities and relations in knowledge graph into low-dimensional vector space. Path-based methods~\cite{PRA,NELL995,pathKgr,MINERVA} start from anchor entities and determine the answer set by traversing the intermediate entities via relational path.
There are also GCN~\cite{gcn} based methods~\cite{GNN1,GNN2} pass message to iterate graph representation for reasoning.

\noindent\textbf{Vision-Language Pre-training}
Vision-language pre-training (VLP) can be divided into three categories based on how they encode images~\cite{empirical}: OD-based region features~\cite{region1,region2,oscar&region3,vilbert&region4,region5,lxmert&region6}, CNN-based grid feature~\cite{veClip,SOHO,pixelbert} and ViT-based patch features~\cite{probing,ALBEF,vilt}. Pre-training objectives are usually: masked language/image modeling (MLM/MIM)~\cite{beit,bert,roberta}, image-text matching (ITM)~\cite{oscar&region3,SOHO,empirical}, and image-text contrastive learning (ITC)~\cite{ALBEF,CLIP,Declip}.

%% file: body/4_method.tex
\section{UKnow}\label{sec:method}
We commence by introducing the overall architecture of \method
in Sec.~\ref{sec:31}. Then the detailed exposition of the data collection process for the new dataset and statistics are presented in Sec.~\ref{sec:da} and Sec.~\ref{sec:Statis}.
In Sec.~\ref{sec:kgm}, we lastly provide the guidance to researchers on how to integrate the multimodal knowledge graph and effectively design a UKnow-based model.

\begin{figure*}[t]
  \begin{center}
   \subfigure[$I_{in}$ and $T_{in}$.]{\label{fig:a}\includegraphics[width=1.0\linewidth]{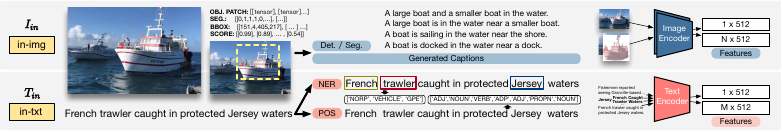}}

   \subfigure[$I_{cross}$, $T_{cross}$ and $IT_{cross}$.]{\label{fig:b}\includegraphics[width=0.49\linewidth]{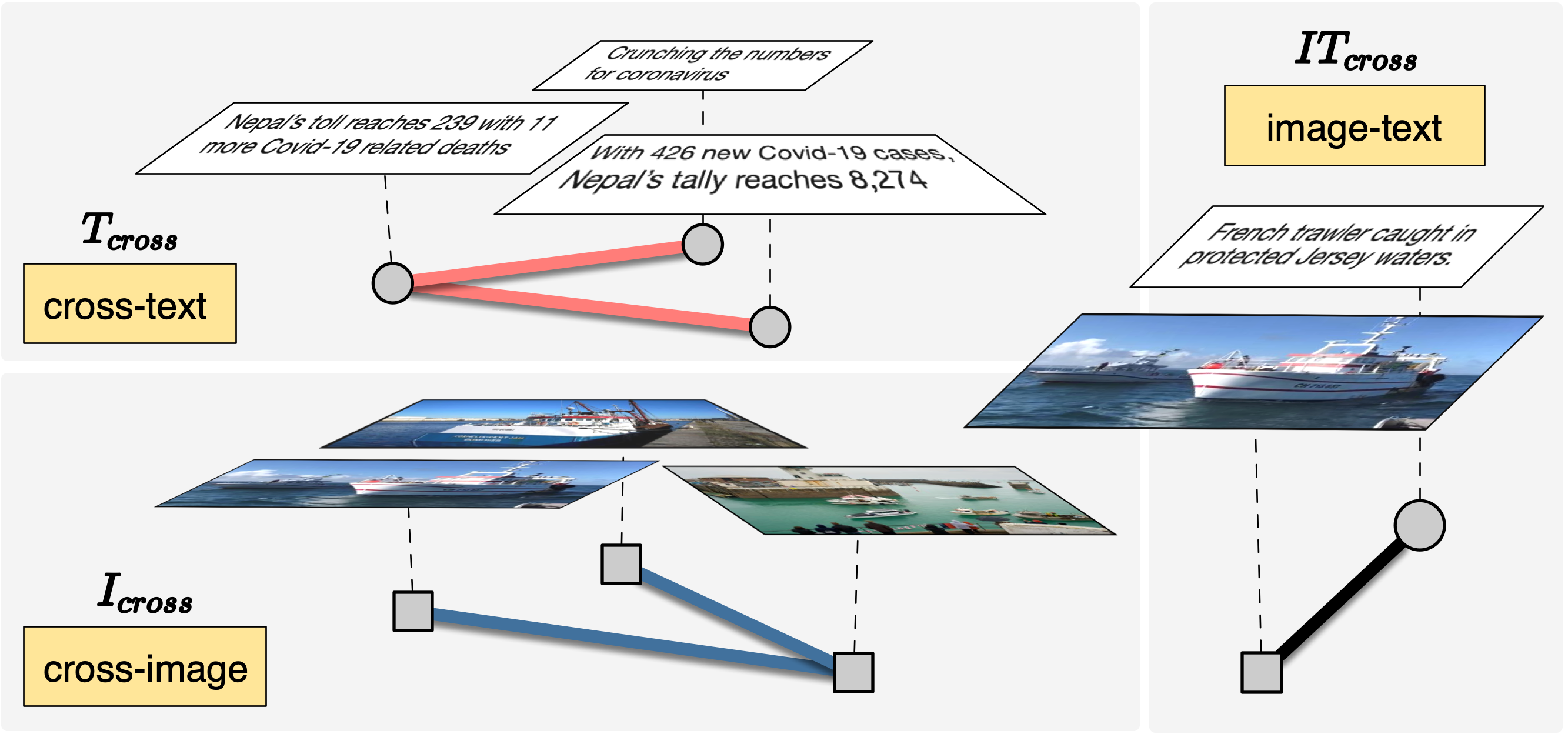}}
    \subfigure[Illustration of the complete UKnow.]{\label{fig:c}\includegraphics[width=0.49\linewidth]{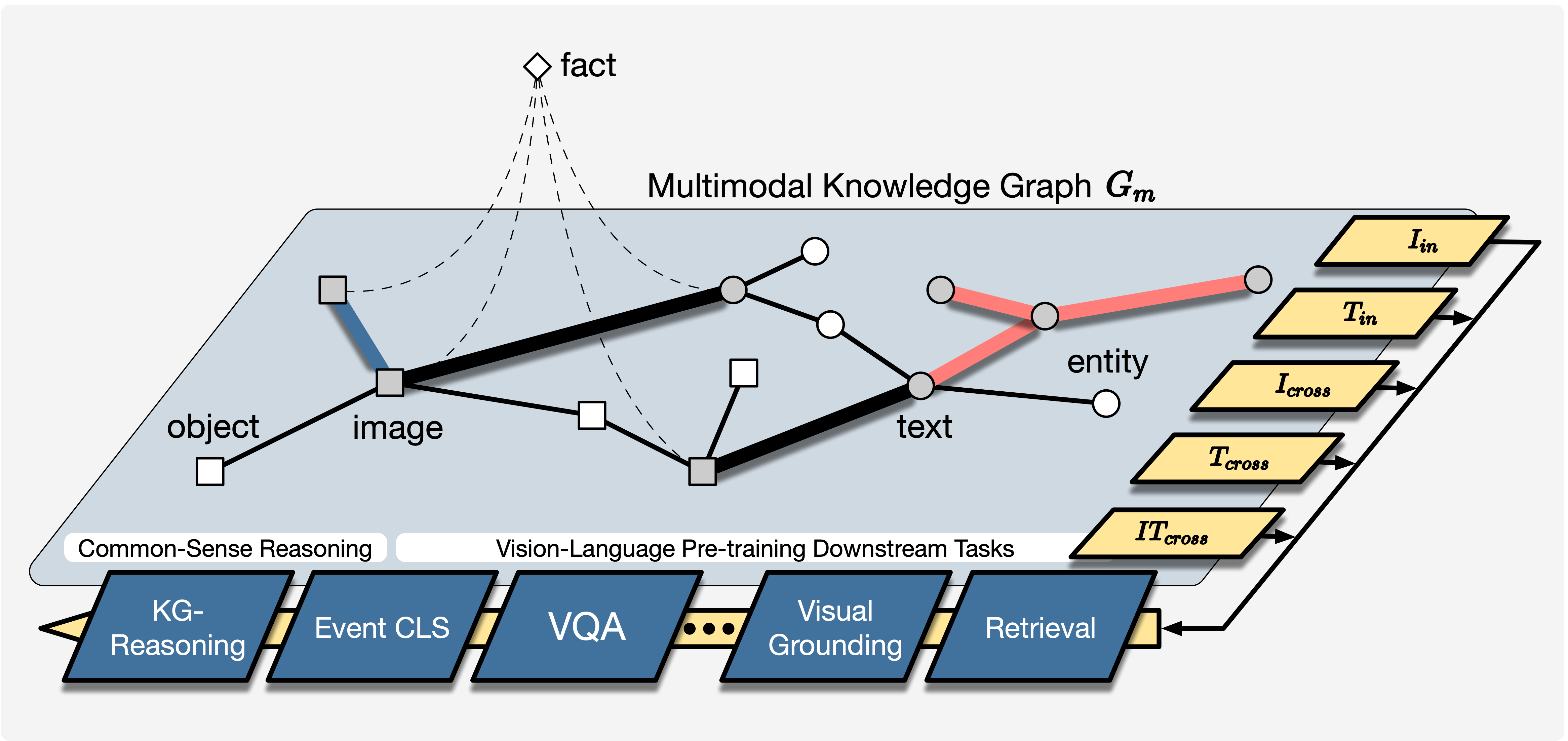}}
    
  \end{center}
  \caption{\textbf{Detailed data organization under \method protocol}, which builds the multimodal (image $\&$ text) graph $\mathbf{G}_m$ based on the \textit{Knowledge-View} ($I_{in}$, $T_{in}$, $I_{cross}$, $T_{cross}$, and $IT_{cross}$). Each node owns up to 22 attributes
	shown as $N_p$ in Fig.~\ref{fig:5}.}
 \vspace{-3mm}
  \label{fig3}
\end{figure*}

Compared to previous libraries-like methods~\cite{liu2016deepfashion,song2021matching} with simple descriptions and meta-information, which lack the logical connection, the most valuable feature of our data processing pipeline is to endow with more logical connections to achieve superior performance in various tasks.
As shown in Fig.~\ref{fig3}, particularly focusing on visual and linguistic modalities, we categorize data knowledge into five unit types. Then we devise an efficient data processing pipeline to help reorganize existing datasets or create a new one under \method format. The construction process of \textit{UKnow} can be invoked separately for any multimodal data to standardize the knowledge. As shown in Fig.~\ref{fig:5}, the whole pipeline is mainly empowered by three parts: \textit{content extraction}, \textit{information symbolization}, and \textit{knowledge construction.}

\subsection{Construction Pipeline for UKnow Protocol}
\label{sec:31}

\begin{figure}[t]
\vspace{-3mm}
    \begin{center}
    \includegraphics[width=0.9\linewidth]{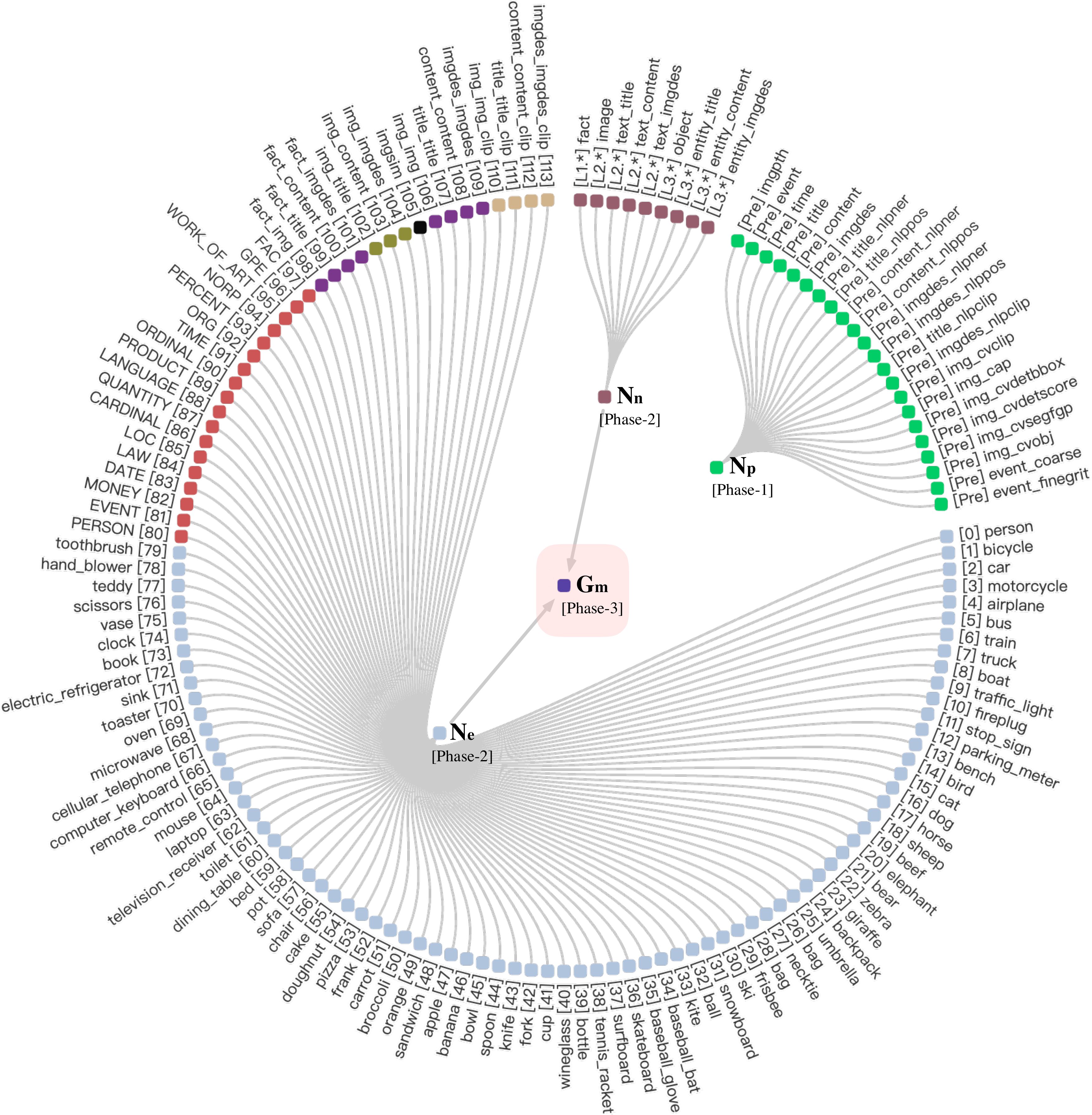}
    \end{center}
  
     \caption{
\textbf{Pipeline of dataset construction following \textit{UKnow} protocol.} Phase-1: \textit{Content Extraction} ($N_p$), Phase-2: \textit{Information Symbolization} ($\mathbf{N}_n$, $\mathbf{N}_e$), and Phase-3: \textit{Knowledge Construction} ($\mathbf{G}_m$). $\mathbf{N}_n$ hides the real node index for easy understanding, the actual number is much more than $\mathbf{N}_e$.
    }
\vspace{-6mm}
\label{fig:5}
\end{figure}

\noindent\textbf{Phase-1: Content Extraction.}
\textit{Content Extraction} is used to extract useful information from different fields by pre-trained deep learning models. The pre-processing functions are designed as $\mathbf{P}=\{P_1, P_2, \ldots, P_k\}$. Note that $\mathbf{P}$ can be replaced / added / disabled freely as needed. We choose pre-trained models with both global descriptions and semantic level granularity:
\begin{wrapfigure}{l}{0.5\textwidth}
\begin{equation}
    \mathbf{P} = \left\{
    \begin{aligned}
    \{P_1,P_2\},& \ \ Image\ Encoder\text{~\cite{CLIP,he2022learn}}\\
    \{P_3,P_4,P_5\},& \ \ Image\ Caption\text{~\cite{cap1,cap2,cap3}}\\
    \{P_6\},& \ \ Image\ Det./Seg.\text{~\cite{detectron2}}\\
    \{P_7\},& \ \ Text\ Encoder\text{~\cite{CLIP}}\\
    \{P_8\},& \ \ Text\ NER/POS\text{~\cite{minilm}}\\
    \{P_9\},& \ \ Annotation\\    
    \end{aligned}
    \right.
\end{equation}
\end{wrapfigure}
where \textit{Det. / Seg.} and \textit{NER / POS} refer to \textit{Detection / Segmentation} and \textit{Named Entity Recognition / Part-of-Speech tagging}. Then we construct the $N_p^{ori} = \mathbf{P}(I,T)$ ($I$ is a RGB-image and $T$ is a text) which contains a wealth of external knowledge. At this stage, all inputs concurrently go through the entire $\mathbf{P}$. It also supports the combined use of pre-trained models such as $P_6\rightarrow P_2$ (\textit{e.g.}, extracting the features of each  object detected from the image). The final output of \textit{Content Extraction} can be formulated as $N_p = Merge (N_p^{ori})$. $Merge$ transforms the original output $N_p^{ori}$ into a K:V dictionary $N_p$. The KEY of $N_p$ are shown in top right corner of Fig.~\ref{fig:5} ($N_p$ [\textit{Phase-1}]). $N_p$ is also used as the attribute of each node in the final output multimodal knowledge graph $\mathbf{G}_m$.

\noindent\textbf{Phase-2: Information Symbolization.}
Since Images and texts cannot be used directly for graph construction, we design the \textit{Phase-2} to number all original or generated data by a certain rule, then \textit{Phase-3} links these nodes to make a multimodal graph.
\textit{Information Symbolization} is used to subscript $N_p$ to edge index $\mathbf{N}_e$ or node index $\mathbf{N}_n$: (1) The symbolization for edges $\mathbf{N}_e$ is based on the category or visual / semantic similarity. For example, ``[$111$] title\_title\_clip'' is a kind of parallelism edge which is constructed by the cosine similarity of clip features of news titles. (2) The symbolization for nodes $\mathbf{N}_n$ is divided into three levels: [fact, image / text, object / entity]. As shown in Fig.~\ref{fig:5}, [$L_1.*$] means fact-level which is an abstraction of a piece of news. The real index used in our multimodal knowledge graph would be $\{L_1.0,L_1.1,L_1.2,...\}$. Similarly, [$L_2.*$] means image / text-level which is the symbolization of images or texts from news, [$L_3.*$] is the object in image or entity in text. The index for all nodes is eventually shuffled, that is, the real index would be $\{L_1.0,L_2.1,L_1.2,L_3.3,L_3.4,...\}$.

\begin{wraptable}{r}{0.5\textwidth}
\vspace{-4.7mm}  
    \tablestyle{5pt}{1.0}
    \def\w{20pt} 
    \setlength\tabcolsep{2pt}
    \caption{\textbf{Edge ($\mathbf{N}_e$) construction and statistics}.
    }     
    \resizebox{\linewidth}{!}{
      \begin{tabular}{ccc|c}
        \shline
        \bf Phrase &\bf Construction Method & \bf View&\bf Num.\\
        \shline
        \multirow{4}{*}{Phrase-2}  &Detection Category & $I_{in}$ & 648,871 \\
        & NER Category & $T_{in}$ & 1,606,936 \\
        & Similarity\&Manual Annotation  & $IT_{cross}$ & 684,207 \\
        & Similarity\&Manual Annotation  & $T_{cross}$,$I_{cross}$ & 140,133 \\
        \hline
        Phrase-3 &Manual Event Annotation & - &593,670 \\
        \shline
      \end{tabular}
      }
      \label{tab:3}
    \vspace{-4mm}  
\end{wraptable}

\noindent\textbf{Phase-3: Knowledge Construction.}
We categorize data knowledge into five unit types, namely, in-text ($T_{in}$), in-image ($I_{in}$), inter-text ($T_{cross}$), inter-image ($I_{cross}$), and image-text ($IT_{cross}$) which are together called \textit{Knowledge-View} detailed in Fig.~\ref{fig:a} and Fig.~\ref{fig:b}.

In this phase, we aggregate two kinds of internal knowledge ($I_{in}, T_{in}$) and three kinds of associative knowledge ($I_{cross}, T_{cross}, IT_{cross}$) in one graph $\mathbf{G}_m$, which are usually introduced independently in previous studies.
\textit{Knowledge Construction} takes as input the edge index $\mathbf{N}_e$ and node index $\mathbf{N}_n$ numbered by \textit{Phase-2} and output the multimodal knowledge graph $\mathbf{G}_m$ (Fig.~\ref{fig:c}). Since $\mathbf{N}_e$ and $\mathbf{N}_n$ are both isolated, we use four kinds of correlation methods including semantic similarity, visual similarity, annotations, and categories to make connections between $\mathbf{N}_n$ by $\mathbf{N}_e$ shown in Tab.~\ref{tab:3}.

\begin{figure}[t]
  \vspace{-5mm}
    \begin{center}
    \includegraphics[width=1.0\linewidth]{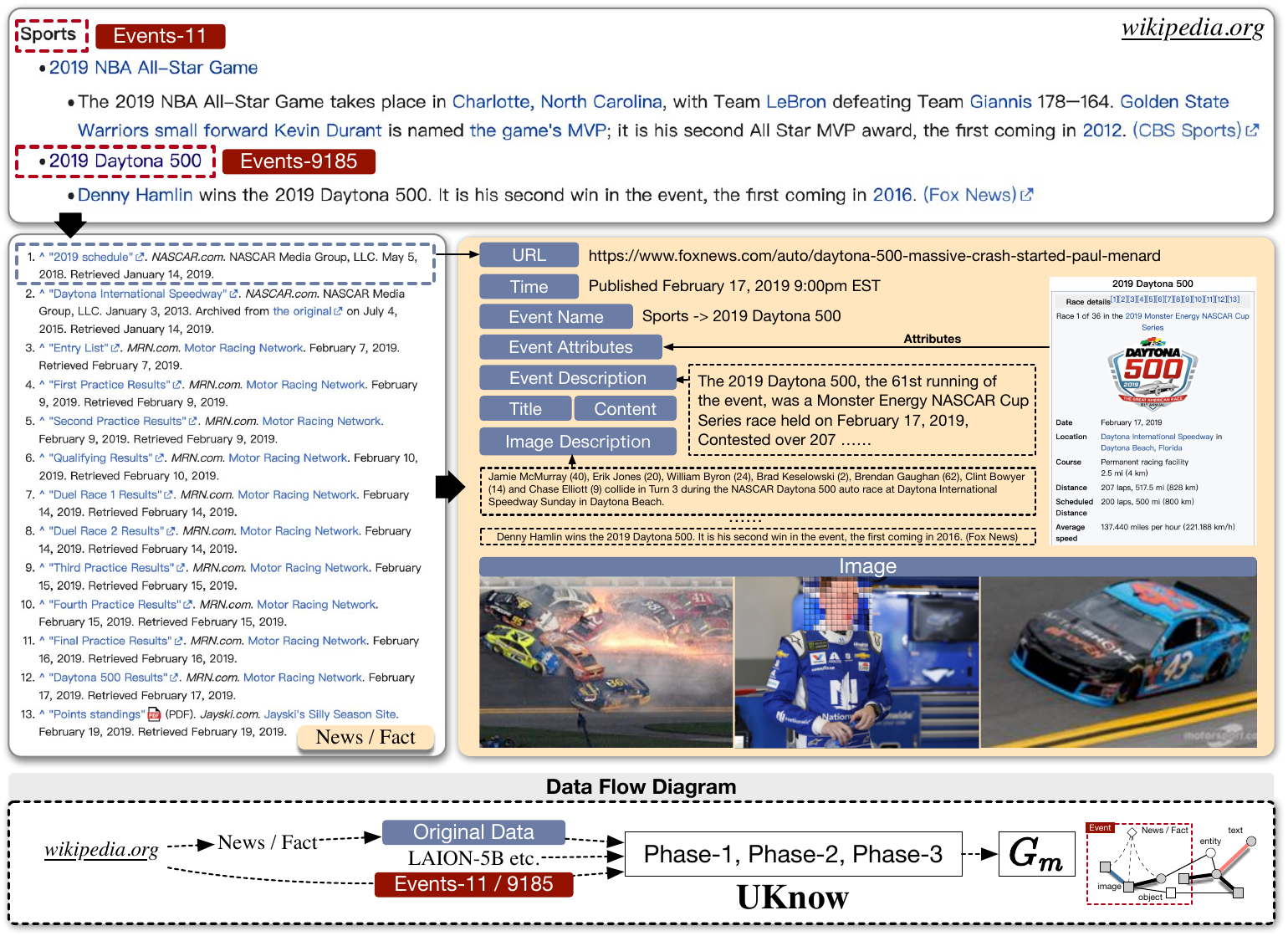}
    \end{center}
  \vspace{-5mm}
     \caption{
\textbf{Event category} labeled on web data and the \textbf{data flow diagram.}}
\label{fig:event}
\end{figure}

\subsection{Dataset Collection}
\label{sec:da}
Following the proposed protocol and three phases, we collect a new dataset, a large-scale multimodal knowledge graph from public international news. Specifically, based on the Wikipedia API~\cite{Wikipedia-API} and our crawler system, we grab all the data of ``\textit{Worldwide Current Events}'' from Wikipedia. 
As demonstrated in top of Fig.~\ref{fig:event}, we propose two category sets of news event called: \textit{Event-11} and \textit{Event-9185}, which is coarse-grained and fine-grained respectively. For example, \textit{``Sports''} is a kind of coarse-grained event label in \textit{Event-11} and \textit{``2019 Daytona 500''} is a fine-grained label in \textit{Event-9185}, detailed in Tab.~\ref{tab:coraevent}. Since Wikipedia only records the news URL (downward black arrow in Fig.~\ref{fig:event}) and the HTML of original news from different news platforms is inconsistent, it is difficult to design a uniform crawler to get the well-structured raw data of news. Thus, we manually read each news and collect the original data (rightward black arrow). By this way, each news in our dataset is marked with extremely clean \textbf{title}, \textbf{content}, \textbf{time}, [\textbf{image}], \textbf{image description}, \textbf{event description}, [hierarchical] \textbf{event name} (\textit{e.g.}, \textit{``Armed conflicts and attacks$\rightarrow$War in Donbass''}), and \textbf{event attribute} (location, date, \textit{etc}). Subsequently, as shown in bottom right of Fig.~\ref{fig:event}, we apply the designed pipeline to sequentially undergo phases 1/2/3 to restructure the above extracted raw data, resulting in the knowledge graph under the \method format.

\begin{table*}[!t]
\tablestyle{5pt}{1.0}
\def\w{20pt} 
\setlength\tabcolsep{2pt}
\caption{\textbf{Details of the event category.}
  \vspace{1mm}
  }
\resizebox{\linewidth}{!}{
  \begin{tabular}{cccc|cccc}
  \bf Event Name (\textit{Event-11}) & \bf Visual & \bf Texual & \bf All  & \bf Event Name (10 examples of \textit{Event-9185}) & \bf Visual & \bf Texual & \bf All \\
  \shline
  Armed conflicts and attacks & 87,346 & 90,157 & 177,503 & Saudi Arabian-led intervention in Yemen & 555 & 258 & 813 \\ %
  Arts and culture & 11,059 & 14,896 & 25,955 & \scalebox{1.2}{\tiny A boat carrying Indonesian migrants capsizes off the southern coast of Malaysia} & 46 & 19 & 65\\ %
  Business and economy & 12,598 & 25,565 & 38,163 & Travel restrictions related to the COVID-19 pandemic & 753 & 796 & 1,549 \\ %
  Disasters and accidents & 28,062 & 47,459 & 75,521 & GameStop short squeeze & 45 & 175 & 220 \\ %
  Health and environment & 230,926 & 258,349 & 489,275 & Opposition to Brexit in the United Kingdom & 383 & 93 & 476 \\ %
  International relations & 37,349 & 56,444 & 93,793 & Gretchen Whitmer kidnapping plot & 167 & 308 & 475 \\ %
  Sports & 15,647 & 31,194 & 46,841 & Legality of euthanasia & 185 & 455 & 640 \\ %
  Law and crime & 69,573 & 86,514 & 156,087 & Ukraine International Airlines Flight 752 (Air Crash) & 314 & 179 & 493 \\ %
  Politics and elections & 74,477 & 72,714 & 147,191 & Manhattan blackout & 269 & 90 & 359 \\ %
  Science and technology & 4,062 & 15,556 & 19,618 & 2019 Lagos school collapse & 524 & 119 & 643 \\ %
  Others & 236 & 184 & 420 & ... & ... & ... & ...\\
  \end{tabular}
  }
  \label{tab:coraevent}
    \vspace{-3mm} 
\end{table*}

\begin{table*}[!t]
    \tablestyle{5pt}{1.0}
    \def\w{20pt} 
    \setlength\tabcolsep{8pt}
    \vspace{-2mm}
     \caption{\textbf{Partition of our dataset.}}
     \vspace{1mm}
    \resizebox{\linewidth}{!}{
      \begin{tabular}{l|cccccccccc}
          \toprule
          \\[-3.2ex]
        \multirow{2}{*}{\bf PARTITION} &\multicolumn{2}{c}{$T_{in}$}  &\multicolumn{2}{c}{$I_{in}$} &\multicolumn{2}{c}{$T_{cross}$}&\multicolumn{2}{c}{$I_{cross}$}& \multicolumn{2}{c}{$IT_{cross}$}\\
        \cline{2-11}\\[-2.4ex] 
        &\bf NODE&\bf EDGE&\bf NODE&\bf EDGE&\bf NODE&\bf EDGE&\bf NODE&\bf EDGE&\bf NODE&\bf EDGE\\
        \hline
        Training Set&448,691 &8,030,531 &501,564 &979,287 &250,858 &396,200 &69,911 &421,628 &765,654 &382,827 \\
        Validation Set& 37,488 & 100,280 & 12,126 & 12,212 & 69,533 & 57,162 & 15,532 & 97,272 & 9,764 & 4,882\\
        Testing Set& 37,668 & 100,375 & 12,182 & 12,261 & 69,286 & 55,464 & 15,336 & 99,303 & 9,622 & 4,811\\
        \hline
        Pre-training Set& 228,339 & 435,659 & 343,458 & 325,755 & 101,880 & 314,918 & 47,017 & 271,593 & 278,058 & 139,029\\
        Fine-tuning Set& 75,924 & 82,350 & 65,809 & 61,850 & 19,185 & 59,832 & 8,880 & 52,772 & 52,522 & 26,261\\
        Testing Set& 34,422 & 28,219 & 22,809 & 22,278 & 6,633 & 21,360 & 3,074 & 17,754 & 18,186 & 9,093\\
        [-0.6ex]
        \bottomrule
     \end{tabular}
  }\vspace{-2mm}
 \label{tab:2}
\end{table*}

Furthermore, in addition to utilizing intricate annotation files (\textit{e.g.}, Fig.~\ref{fig:event}) as inputs mentioned above, another major advantage of the proposed conversion pipeline is its ability to accommodate common image-text pair annotations expressed in the format of \textit{``[image description] \textbackslash t ./xxx.jpg \textbackslash n''}), as the fundamental input.
This design allows \method to automatically construct a new dataset with more useful information from an existing image-text pair dataset. Taking LAION-5B~\cite{schuhmann2022laion} as an example, which solely comprises pairs of images and text, our pipeline can extract more features from them like objects, and thus expand LAION-5B into a larger and more practical dataset.
However, given the absence of high-level event logic, this type of input does not lend itself to the creation of [$L_1.*$] nodes and event-related edges.

\subsection{Dataset Statistics and Visualizations}
\label{sec:Statis}
Through data collection and processing in Sec.~\ref{sec:da}, our dataset comprises 1,388,568 nodes, of which 571,791 are relevant to vision (i.e., pertaining to a news image or a visual object), and 3,673,817 triples. %
The partitioning of our dataset is presented in Tab.~\ref{tab:2}, with all partitions being randomly sampled. Moreover, as depicted in Fig.~\ref{fig:his}, we present the histogram of all indices in \textit{UKnow}. Considering our dataset is a multimodal knowledge graph, \textit{i.e.}, each node corresponds to a multimodal data, and each edge serves the purpose of connecting either single or cross-modal nodes.

\begin{table}[t]
	\centering
	\caption{\textbf{Histogram of the number of indexes in our dataset}. The x-axis in the upper left corner (Node Index) corresponds to the order of the $I_n$ in Fig.~\ref{fig:5}.}
		\includegraphics[width=0.87\linewidth]{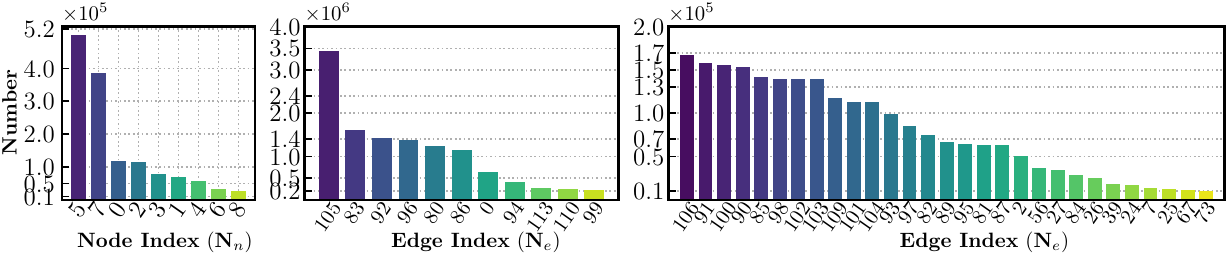}
	\label{fig:his} 
 \vspace{-5mm}  
\end{table}

The top-2 number of nodes are ``[$L_3.*$] objects'' (501,880) and ``[$L_3.*$] entity\_content'' (386,561) which belong to $I_{in}$ and $T_{in}$ respectively.
The former represents visual objects extracted from images, and the latter means text entities extracted from news contents. 
The maximum number of edges is ``[$105$] imgsim'' (3,447,990) which is a kind of associative knowledge from $I_{cross}$.

\begin{wraptable}{r}{0.6\textwidth}
\vspace{-2mm}
	\begin{center}
	\caption{\textbf{Histogram of the number of other edge indexes.}}
        \includegraphics[width=0.9\linewidth]{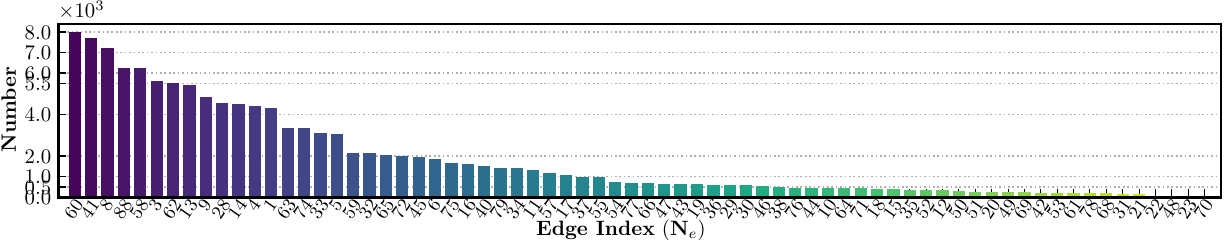}
	\end{center}
        \begin{center}
	\caption{\textbf{The variation in different similarity thresholds.}}
        \includegraphics[width=0.9\linewidth]{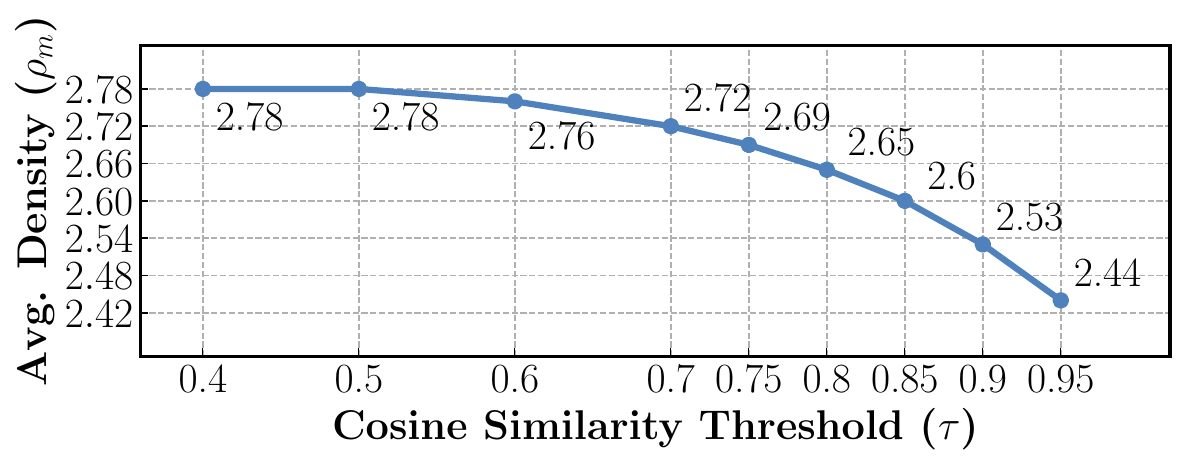}
	\end{center}
	\label{fig:tabvar}
\tablestyle{5pt}{1}
\setlength\tabcolsep{0.8pt}
\def\w{20pt} 
\caption{\textbf{The graph density and mainstream node types.}}
\resizebox{\linewidth}{!}{
    \begin{tabular}{c|c|cccc|cccccc}
    \textbf{$\tau$} & \textbf{Density ($\rho$)} & \scriptsize 0,1 &\scriptsize 2,3  &\scriptsize 4,5   &\scriptsize 6,7 &\scriptsize 8,9 & \scriptsize10,11 & \scriptsize12,13 &\scriptsize 14,15&\scriptsize 16,17&\scriptsize $\geqslant$18 \\ [-0.2ex]
    \shline 
   \cellcolor[HTML]{EFEFEF} & \textbf{NodeNum.} & 615k & 463k  & 132k  & 71k  & 55k  & 32k  & 14k  &4k  &703 &78 \\
    [-0.7ex]
    \cellcolor[HTML]{EFEFEF} \multirow{-2}*{0.8} & \textbf{MainType}   & Entity & Object  & Title  & Image  & 
    \multicolumn{6}{c}{Content}\\
    \end{tabular}%
  }
	\label{tab:rb2}  
 \vspace{-2mm}
\end{wraptable}
Fig.~\ref{fig:tabvar} shows the variation in different thresholds. $\tau$ indicates the threshold of cosine similarity which controls whether edges are built between nodes. 
It can be adjusted according to needs (\textit{e.g.}, storage, computational complexity, fineness). The default setting of $\tau$ is \hl{0.8}. $\rho_m$ indicates the average edge number of connections per node in the entire graph, which demonstrates the density of a graph.
As shown in Tab.~\ref{tab:rb2}, the whole graph is sparse with \textit{ENTITY} as the main nodes of the background, and the subject element of the dense region is \textit{CONTENT}.
$\rho$ means the number of edges, \textit{i.e.}, there are 615k nodes with 0-edge or 1-edge, 463k nodes with 2-edges or 3-edges, and so on. 
The Mean Density ($\rho_m$) in Fig.~\ref{fig:tabvar} is calculated as a weighted average of $\rho$ and the number of nodes.

Tab.~\ref{tab:coraevent} shows all the categories in \textit{Event-11}, and 10 examples in \textit{Event-9185}. ``Visual'' means the number of nodes belonging to images or objects. ``All'' means the number of all nodes marked with this event category. Generally speaking, \textit{Event-9185} is specific to an exact human activity and can be used to learn the semantic relevance of news contents, while \textit{Event-11} is more like a categorization of news events, which is benefit for archiving news materials through a trained 
classification model.

\section{Usage of UKnow}
\label{sec:kgm}
\subsection{UKnow for Common-sense Reasoning}
Since \textit{UKnow} is reasoning compatible, \textit{i.e.}, it naturally supports all KG-reasoning models, we directly implemented the commonly used KG-reasoning models (\textit{e.g.}, TransE~\cite{TransE}, Q2B~\cite{Q2B}) on \textit{UKnow}. We propose a plug-in module which aggregates node features within a small sub-graph region to achieve a better central node features. We briefly introduce how to implement this module. Suppose $N(e)\equiv\{e_{neib}|r(e_{neib},e)\vee r(e,e_{neib}),r\in\mathcal{R}\}$ is the collection of neighbors of each central node $e$. The calculation expression of the new representation ${\rm e}^\prime$ of ${\rm e}$ is as follow:
\begin{equation}
    {\rm e}^\prime = {\rm MLP}(Flatten({\rm ReLU}(\omega_n \star ({\tau}^\prime ({\rm e}, {N_{\rm e}}^\prime)) + {\rm b}_n)), \label{}
\end{equation}
where ${\rm e} \in \mathbbm{R}^d $ is the node feature before enhancement, ${\rm e}^\prime $ is the new feature, $\star$ denotes a 2D convolution operation, $\omega_n$ is the filter, ${\rm b}_n $ is the bias and the specification of ${\rm MLP}$ is $\mathbbm{R}^{m_1 \times m_2} \times \mathbbm{R}^d$. The concat function ${\tau}^\prime ({\rm e}, N_{\rm e}) \in \mathbbm{R}^{m_1 \times m_2}$ as $[{\rm e}; {{\rm e}_{neib}}^{1^\prime} ; {{\rm e}_{neib}}^{2^\prime}; \ldots ;{{\rm e}_{neib}}^m]$ where ${{\rm e}_{neib}}^i \in {N_{\rm e}^\prime}$.

\subsection{UKnow for Vision-Language Pre-training}
Following the recent works~\cite{clipEvent}, our work applies CLIP~\cite{CLIP} as the pre-trained backbone benefit from its strong downstream performance. Specifically, the text encoder first tokenize the input text description into the word sequence, and then projects them into word embeddings $\mathbf{W}_0 = \{\mathbf{w}^1_0, \mathbf{w}^2_0, \cdots, \mathbf{w}^N_0\} \in \mathbb{R}^{N \times d^t}$. $\mathbf{W}_0$ is fed into a $L$-layer Transformer~\cite{Vaswani:Transformer} with the architecture modifications described in BERT~\cite{bert}. And the final text embedding $\mathbf{z}^T$ is obtained by projecting the last token, which corresponds to the \texttt{[EOS]} (the end of sequence) token, from the last layer of the text encoder, \textit{i.e.}, $\mathbf{z}^T = \texttt{TextProj}(\mathbf{w}^N_L),\mathbf{z}^T \in \mathbb{R}^d$. As for the vision encoder, the input image $I$ is first split into $M$ non-overlapping patches, and projected into a sequence of patch tokens $\mathbf{E_0} \in \mathbb{R}^{M \times d^v}$.
Then, $\mathbf{E_0}$ is fed into a $L$-layer Transformer-based architecture along with a learnable \texttt{[CLS]} token $\mathbf{c}_0$. The final image embedding $\mathbf{z}^I$ is obtained by projecting the \texttt{[CLS]} token from the last layer of the vision encoder, \textit{i.e.}, $\mathbf{z}^I = \texttt{VisProj}(\mathbf{c}^v_L,\mathbf{E}^v_L)), \mathbf{z}^I \in \mathbb{R}^d$. Since we have \textit{Knowledge-View}, a new dimension $\mathbf{z}^k$ which is used to represent knowledge is introduced:
\begin{align}
\mathbf{z}^k = \texttt{Concat}(I_{in}(\mathbf{z}^I),T_{in}(\mathbf{z}^T),I_{cross}(\mathbf{z}^I),T_{cross}(\mathbf{z}^T)),
\end{align}
where $I_{in}(\cdot)$ and $T_{in}(\cdot)$ mean to get the embedding of the [$L_3.*$] nodes ($\mathbf{N}_n$) from $\mathbf{G}_m$ via $\mathbf{N}_e$, $I_{cross}(\cdot)$ and $T_{cross}(\cdot)$ mean to get the embedding of [$L_2.*$] from $\mathbf{G}_m$. Therefore, the similarity score between the image, text and knowledge can be calculated with the cosine similarity as follow:
\begin{align}
s(T, I, k) =  \frac{{\mathbf{z}^T}^{\top} \mathbf{z}^I}{\| \mathbf{z}^T \| \| \mathbf{z}^I \| } + \frac{{\mathbf{z}^k}^{\top} \mathbf{z}^I}{\| \mathbf{z}^k \| \| \mathbf{z}^I \| } + \frac{{\mathbf{z}^k}^{\top} \mathbf{z}^T}{\| \mathbf{z}^k \| \| \mathbf{z}^T \| }.
\end{align}

%% file: body/5_exp.tex
\subsection{UKnow Baseline} \label{sec:exp}

Upgrading AI from understanding objects (\textit{e.g.}, an apple) as in most current vision tasks to understanding complex human activities (\textit{e.g.}, an event), to understanding the logic between entities or objects, and to achieving higher-order intelligence, is always the thing we would like to pioneer.
Thus, in this section, we naturally present a series of novel logic-rich downstream tasks as the baselines for our dataset. Specifically, Common-sense Reasoning is a conventional and fundamental task in our domain, aligning closely with our dataset. Then we perform multiple downstream tasks to verify the performance of the pretrained model trained with our dataset. For more details about task description/training setting/evaluation metric/analysis, please refer to Sec.~\ref{sec:training_detail}.

\noindent\textbf{Common-sense Reasoning.} 
We implement the Q2B$^*$ with our \method based plug-in module based on Q2B~\cite{Q2B} and BETAE$^*$ based on BETAE~\cite{BetaE_KGreasoning}. 
As shown in Tab.~\ref{tab:4}, BETAE$^*$ achieves on average \textbf{21.64\%} and \textbf{21.23\%} MRR on the validation and testing set of our dataset. It indicates that our \method based module can significantly improve the performance of existing methods.

\noindent\textbf{Multimodal Event Classification.} 
As shown in Tab.~\ref{tab:tab6}, TCL~\cite{TCL} achieves on \textbf{66.80\%} and \textbf{55.87\%} on ACC@1 when using the image-input on the \textit{Event-11} and \textit{Event-9185}. respectively. 
We add a late-fusion module after the image/text encoder for all methods to support multimodal classification. 
Results show that TCL obtains gains of \textbf{1.89\%} and \textbf{5.02\%} compared with the singlemodal input, which demonstrates that multimodal pre-training is more helpful for downstream multimodal tasks.

\noindent\textbf{Single- \& Cross-Modal Retrieval.}
As shown in Tab.~\ref{tab:tab5}, TCL~\cite{TCL} achieves on \textbf{33.24\%}, \textbf{43.37\%} and \textbf{45.22\%} R@1, R@5, R@10 on the zero-shot setting of image retrieval. The results are \textbf{58.89\%}, \textbf{68.47\%} and \textbf{73.91\%} when fine-tuning the pre-trained parameters, which means the pre-training$\rightarrow$fine-tuning strategy is extremely beneficial for downstream retrieval.

\noindent\textbf{Visual Task Adaptation.}
\label{sec:pdtask}
As shown in Tab.~\ref{tab:tab7}, our approach obtains gains of avg. \textbf{1.14\%} compared with the origin CLIP when fairly using the same \textit{UKnow}'s data for the upstream pre-training. 
It is essential to highlight that the image-text PAIR constitutes only one type of data in our protocol. By leveraging the capabilities of \textit{UKnow}, our pre-trained CLIP model can effectively comprehend the inherent knowledge, resulting in superior performance than original CLIP model (Tab.~\ref{tab:tab7}, Row2).

%% file: body/6_con.tex
\section{Conclusion}\label{sec:conclusion}
This paper presents a unified knowledge protocol called \method to establish the standard of knowledge from the perspective of data. 
Following this protocol, we collect a novel and the largest multimodal knowledge graph dataset from public international news with rich news event annotations, which can help intelligent machines understand human activities and history.
The specific tasks addressed in this paper are the common-sense reasoning and vision-language pre-training. The former is a typical task in the knowledge graph field, and the latter brings knowledge to various downstream tasks.
We also present a series of novel logic-rich downstream tasks to showcase the advantages of \textit{UKnow}.
In future work, we will continuously expand the data of different scales based on the \method protocol.

%% file: body/7_app.tex
\section{Addition Statement for Our New Dataset}
\label{sec: addotional_data}

\subsection{Dataset Documentation and Intended Use}
We offer a detailed overview of our dataset statistics in Sec.~\ref{sec:Statis}. To facilitate better understanding and ease of access, we have made our dataset project available on ModelScope at: \textit{https://www.modelscope.cn/datasets/
yutong/UKnow/summary}, which includes \href{https://www.modelscope.cn/datasets/yutong/UKnow/summary}{dataset summary}, \href{https://www.modelscope.cn/datasets/yutong/UKnow/dataPeview}{data preview}, \href{https://www.modelscope.cn/datasets/yutong/UKnow/quickstart}{quickstart} and \href{https://www.modelscope.cn/datasets/yutong/UKnow/files}{data files}. 

The detailed data organization and corresponding download links are listed below:
\begin{itemize}[labelsep=0.4em, leftmargin=1em,itemindent=0em]
    \vspace{-8pt}
    \setlength{\itemsep}{0pt}
    \setlength{\parsep}{0pt}
    \setlength{\parskip}{0pt}
\item Original data: We gather our data from publicly available international news sources, accumulating a substantial volume of images and text. Subsequently, we compress the collected data into several zip archives and store them in \href{https://www.modelscope.cn/datasets/yutong/UKnow/files}{original\_data}: \underline{\texttt{UKnow/raw\_data/*}}.
\item Processed data:
    \begin{itemize}[label=\textbullet, leftmargin=1em,itemindent=0em]
        \setlength{\itemsep}{0pt}
        \setlength{\parsep}{0pt}
        \setlength{\parskip}{0pt}
        \item Pre-node $N_p$: Building upon Phase-1, we leverage pre-trained deep learning models to extract valuable information from various domains. The resultant output from Phase-1 is structured as a dictionary and is then stored and saved to \href{https://www.modelscope.cn/datasets/yutong/UKnow/file/view/master?fileName=processed_data/pre_node0.npy}{pre\_node}:
        \underline{\texttt{UKnow/processed\_data/pre\_node*}}.
        \item Node index $N_n$ and Edge index $N_e$: As the outcomes acquired in Phase-1 (e.g., $N_p$) are not directly applicable for graph construction, we employ an information symbolization strategy to organize them into indices, namely $N_n$ and $N_e$, which are subsequently saved to \href{https://www.modelscope.cn/datasets/yutong/UKnow/file/view/master?fileName=processed_data/edge_index.pickle}{index}: \\
        \underline{\texttt{UKnow/processed\_data/*\_index*.pickle}}.
        \item Knowledge graph $G_m$: Finally, we consolidate two types of internal knowledge ($I_{in}, T_{in}$) and three types of associative knowledge ($I_{cross}, T_{cross}, IT_{cross}$) into into one knowledge graph ($\mathbf{G}_m$), which is stored as a dictionary in \href{https://www.modelscope.cn/datasets/yutong/UKnow/file/view/master?fileName=processed_data/graph.pickle}{graph}: 
        \underline{\texttt{UKnow/processed\_data/graph*.pickle}}.
    \end{itemize}
\end{itemize} 

Our dataset is intended for academic use and the corresponding license is based on: https://www.contributor-covenant.org/zh-cn/version/1/4/code-of-conduct.html, which was created by Coraline Ada Ehmke in 2014 and is released under the \href{https://creativecommons.org/licenses/by-nc-nd/4.0/deed.en}{CC BY-NC-ND 4.0}. 

\subsection{Author statement}
We confirm the data licenses and that we bear all responsibility in case of violation of rights.

\subsection{Hosting, licensing, and maintenance plan}
\paragraph{Hosting and Licensing.} Our dataset is hosted on ModelScope. Moreover, we furnish the relevant licenses in accordance with ModelScope at: https://www.contributor-covenant.org/zh-cn/version/1/4/code-of-conduct.html, which was created by Coraline Ada Ehmke in 2014 and is released under the \href{https://github.com/EthicalSource/contributor_covenant/blob/release/LICENSE.md}{CC BY 4.0 License}.

\paragraph{Introduction to ModelScope.} ModelScope is a platform designed for managing and optimizing machine learning models. It provides various tools and features to streamline the model development process, including version control, performance monitoring, and collaboration capabilities. As for managing datasets, ModelScope offers robust functionality for organizing, storing, and accessing data. Users can upload datasets to the platform, where they are securely stored and can be easily accessed by authorized team members. ModelScope also supports versioning of datasets, allowing users to track changes over time and ensure reproducibility in their experiments. Additionally, the platform provides tools for data preprocessing, visualization, and analysis, helping users to efficiently prepare their data for model training and evaluation. Overall, ModelScope offers comprehensive support for managing datasets throughout the machine learning lifecycle. Therefore, we choose ModelScope as our hosting platform.

\paragraph{Usage of ModelScope.} To enable users to directly utilize all models on the ModelScope platform without configuring the environment, ModelScope integrates an online Notebook programming environment on its website and offers official mirrors for developers. These official mirrors allow users to bypass all installation and configuration steps, providing immediate access to the models. Currently the latest version of the CPU mirror and GPU mirror can be obtained from the office \href{https://github.com/modelscope/modelscope\#docker}{ModelScope repository}.

Users also can setup local python environment using following commands:
\lstdefinestyle{mystyle}{
    backgroundcolor=\color{lightgray!20},   
    basicstyle=\scriptsize\ttfamily,       
    keywordstyle=\color{blue},               
    commentstyle=\color{green},              
    stringstyle=\color{red},                 
    numbers=left,                            
    numberstyle=\tiny\color{gray},           
    stepnumber=1,                           
    numbersep=5pt,                          
    frame=single,                           
    rulecolor=\color{black},                 
    breaklines=true,                        
    lineskip=0pt,                           
}
\lstset{style=mystyle}  
\lstset{language=Bash}
\begin{lstlisting}
conda create -n modelscope python=3.8
conda activate modelscope
pip install modelscope
\end{lstlisting}
%
%
Then, users can access and enjoy our dataset by:
\lstdefinestyle{mystyle}{
    backgroundcolor=\color{lightgray!20},   
    basicstyle=\scriptsize\ttfamily,       
    keywordstyle=\color{blue},               
    commentstyle=\color{green},              
    stringstyle=\color{red},                 
    numbers=left,                            
    numberstyle=\tiny\color{gray},           
    stepnumber=1,                           
    numbersep=5pt,                          
    frame=single,                           
    rulecolor=\color{black},                 
    breaklines=true,                        
    lineskip=0pt,                           
}
\lstset{style=mystyle}  
\lstset{language=Python}
\begin{lstlisting}
from modelscope.msdatasets import MsDataset
ds =  MsDataset.load(`yutong/UKnow', subset_name=`default', split=`train')
\end{lstlisting}
%

Besides, we strongly recommend that users read the \href{https://www.modelscope.cn/docs}{official documents} for optimal use.

\paragraph{Maintenance Plan.} In future work, we will persistently augment the dataset across various scales following the \method protocol. This endeavor aims to furnish a comprehensive, diverse, and resilient multimodal knowledge graph, thereby facilitating subsequent research endeavors.

\section{Preliminaries}
\noindent\textbf{Multimodal Knowledge Graph.}
An intuitive interpretation of multimodal knowledge graph is that the ordinary knowledge graph only consists of
$<$head, relation, tail$>$ triples like
$<$(``Jony''), Citizen, (``New York'')$>$
, but the multimodal knowledge graph consists of the following:\\
{\footnotesize $<$(``$\mathtt{Jony}$''), $\mathtt{Citizen}$, (``$\mathtt{New York}$'')$>$,}\\
{\footnotesize $<$(``$\mathtt{Jony}$''), $\mathtt{Appearance}$, (``$\mathtt{[Face]}$'')$>$,}\\
{\footnotesize $<$(``$\mathtt{New York}$''), $\mathtt{Landmark}$, (``$\mathtt{[Statue of liberty]}$'')$>$,}\\
{\footnotesize $<$(``$\mathtt{[Air Force One]}$''), $\mathtt{Similarity}$, (``$\mathtt{[Air Force Two]}$'')$>$,}\\
where $(\cdot)$ means a text node and $[\cdot]$ means a image node. The machine cannot understand what \textit{``An old man with white hair''} is without establishing the connection between each word and its physical world meaning. However, with the help of multimodal knowledge graph, as a simple example, it is possible to generate a more informative entity-level sentence (\textit{e.g.}, \textit{``Biden is making a speech''}) instead of a vague concept-level description (\textit{e.g.}, \textit{``An old man with white hair is making a speech''}). To evaluate the effectiveness of multimodal knowledge graph (MMKG), several downstream tasks are often performed on the MMKGs, including common-sense reasoning, vision-language pre-training.

\noindent\textbf{Common-sense Reasoning.}
Common-sense reasoning means answering queries by logic permutations. The specific task in this work is the link prediction. In the inference phase, feeding $<$("America"), Capital$>$ to a reasoning model, the output should be $<$("Washington")$>$.
Various works~\cite{TransE,ComplEx,RotatE,ConvE,JointE,AcrE}
achieve reasoning by embedding entities and relations in knowledge graph into low-dimensional vector space.
For instance,
GQE~\cite{GQE} encodes queries through a computation graph with relational projection and conjunction ($\wedge$) as operators. Path-based methods~\cite{PRA,NELL995,pathKgr,MINERVA} start from anchor entities and determine the answer set by traversing the intermediate entities via relational path.
There are also GCN~\cite{gcn} based methods~\cite{GNN1,GNN2} pass message to iterate graph representation for reasoning. Common-sense reasoning is an extremely popular task in the field of knowledge graph. Since our dataset is based on the knowledge graph, the performance validation on common-sense reasoning is indispensable.

\noindent\textbf{Vision-Language Pre-training}
Vision-language pre-training (VLP) can be divided into three categories based on how they encode images~\cite{empirical}: OD-based region features~\cite{region1,region2,oscar&region3,vilbert&region4,region5,lxmert&region6}, CNN-based grid feature~\cite{veClip,SOHO,pixelbert} and ViT-based patch features~\cite{probing,ALBEF,vilt}. Pre-training objectives are usually: masked language/image modeling (MLM/MIM)~\cite{beit,bert,roberta}, image-text matching (ITM)~\cite{oscar&region3,SOHO,empirical}, and image-text contrastive learning (ITC)~\cite{ALBEF,CLIP,Declip}. In this work, we concentrate on the study of the how to introduce our UKnow into ITC method based on ViT-based patch features.

\noindent\textbf{Image-Text Contrastive Learning.} The recent CLIP~\cite{CLIP} and ALIGN~\cite{align} perform pre-training using a crossmodal contrastive loss on millions of image-text pairs, which achieves remarkable performance on various downstream tasks~\cite{cap1,veClip,dsclip}. MDETR~\cite{mdetr} trains on multi-modal datasets which have explicit alignment between phrases and objects. GLIP~\cite{glip} generates grounding boxes in a self-training fashion, and makes the learned representations semantic-rich. We implement these mainstream methods on our dataset, and also design a basic knowledge-based ITC method with UKnow.

\section{Experimental Details} \label{sec:training_detail}

In this section, we give more details about the computation complexity, training, fine-tuning hyperparameters and evaluation for reference.

\subsection{Common-sense Reasoning}
\noindent\textbf{Datasets.}
Since our dataset is a knowledge graph, we benchmark the performance of KG-reasoning models on our dataset by completing KG-triples. The partitioning of the dataset is illustrated in the upper segment of Tab.~\ref{tab:2}.

\noindent\textbf{Evaluation.}
The specific task of common-sense reasoning in this work is the link prediction. Given a test query $q$ (\textit{e.g.},, $<$(``Jony''), Citizen, (?)$>$), we are interested in discovering non-trivial answers (\textit{e.g.},, ``New York''). That is, answer entities where at least one edge needs to be imputed in order to create an answer path to that entity. Each entity in our multimodal knowledge graph is not limited to a text entity but a multimodal node.
Following~\cite{BetaE&KGreasoning}, for each non-trivial answer $t$ of test query $q$, we rank it against non-answer entities $\mathcal{E} \backslash [\![q]\!]_{\rm test}$~\cite{TransE}. Then the rank of each answer is labeled as $r$. We use Mean Reciprocal Rank(MRR): $\frac{1}{r}$ and Hits-at-$N \ ({\rm \textbf{H}}@N): 1[r \leq N]$ as quantitative metrics.

\begin{table*}[!t]
\tablestyle{5pt}{1.0}
\def\w{20pt} 
  \caption{\textbf{A new benchmark of the common-sense reasoning task.} We report four metrics of each model on the validation and test sets. All experiments were repeated five times and the variance is shown in the table.
  }
\resizebox{\linewidth}{!}{
  \begin{tabular}{ccccc|cccc}
    \bf Model & \bf Val-H@1 & \bf Val-H@3 & \bf Val-H@10 & \bf Val-MRR & \bf Test-H@1 & \bf Test-H@3 & \bf Test-H@10 & \bf Test-MRR\\
    \shline
    TransE~\cite{TransE} & 11.75 $\pm$ 0.113 & 29.04 $\pm$ 0.112 & 31.76 $\pm$ 0.143 & 14.77 $\pm$ 0.153 & 11.26 $\pm$ 0.114 & 21.68 $\pm$ 0.115 & 31.57 $\pm$ 0.127 & 14.66 $\pm$ 0.123 \\
    Q2B~\cite{Q2B} & 14.99 $\pm$ 0.118 & 25.78 $\pm$ 0.135 & 36.76 $\pm$ 0.169 & 18.80 $\pm$ 0.166 & 14.48 $\pm$ 0.119 & 25.17 $\pm$ 0.135 & 36.32 $\pm$ 0.163 & 18.46 $\pm$ 0.134 \\
    Q2B$^*$ & 16.84 $\pm$ 0.115 & 29.00 $\pm$ 0.166 & 38.85 $\pm$ 0.169 & 19.66 $\pm$ 0.158 & 16.35 $\pm$ 0.122 & 28.67 $\pm$ 0.174 & 38.45 $\pm$ 0.184 & 19.27 $\pm$ 0.146 \\\hline    
    BETAE~\cite{BetaE_KGreasoning} & 18.04 $\pm$ 0.129 & 33.02 $\pm$ 0.161 & 41.97 $\pm$ 0.179 & 21.16 $\pm$ 0.167 & 17.65 $\pm$ 0.129 & 32.75 $\pm$ 0.160 & 41.67 $\pm$ 0.177 & 20.75 $\pm$ 0.140 \\
    BETAE$^*$ & 19.02 $\pm$ 0.125 & 33.97 $\pm$ 0.173 & 43.17 $\pm$ 0.199 & 21.64 $\pm$ 0.173 & 18.22 $\pm$ 0.135 & 33.52 $\pm$ 0.187 & 42.68 $\pm$ 0.198 & 21.23 $\pm$ 0.154 \\

    \hline    
    QA-GNN~\cite{yasunaga2021qa} & 18.04 $\pm$ 0.129 & 33.02 $\pm$ 0.161 & 41.97 $\pm$ 0.179 & 21.16 $\pm$ 0.167 & 17.65 $\pm$ 0.129 & 32.75 $\pm$ 0.160 & 41.67 $\pm$ 0.177 & 20.75 $\pm$ 0.140 \\
    QA-GNN$^*$ & 19.02 $\pm$ 0.125 & 33.97 $\pm$ 0.173 & 43.17 $\pm$ 0.199 & 21.64 $\pm$ 0.173 & 18.22 $\pm$ 0.135 & 33.52 $\pm$ 0.187 & 42.68 $\pm$ 0.198 & 21.23 $\pm$ 0.154 \\
    
  \end{tabular}
  }
  \label{tab:4}
\end{table*}

\begin{table*}[!t]
    \tablestyle{5pt}{1.0}
    \def\w{20pt} 
    \setlength\tabcolsep{2pt}
    \caption{\textbf{A new benchmark of the novel event classification task.} All models are fine-tuned in the training set.
    \vspace{-4mm}
    }     
    \scalebox{1.0}{
      \begin{tabular}{lcccc|cc}
        \bf \multirow{2}{*}{Model} & \bf \multirow{2}{*}{IMG} & \bf \multirow{2}{*}{TXT} & \multicolumn{2}{c}{\bf Event-11} & \multicolumn{2}{c}{\bf Event-9185}\\
        & & & \bf ACC@1 & \bf ACC@5 & \bf ACC@1 & \bf ACC@5 \\
    \shline
    CLIP~\cite{CLIP} & $\checkmark$ &  & 65.77 & 76.82 & 54.62 & 63.19\\
    DeCLIP~\cite{Declip} & $\checkmark$ &  &66.43&78.32&54.86&63.82\\
    ALBEF~\cite{ALBEF} & $\checkmark$ &  &66.29&77.84&55.03&63.47\\
    TCL~\cite{TCL} & $\checkmark$ && 66.80& 78.91& 55.87& 64.33\\
    \hline
    CLIP && $\checkmark$  & 64.32 & 75.92 & 57.48 & 65.78\\
    DeCLIP && $\checkmark$  &65.89&77.51&59.76&67.81\\
    ALBEF && $\checkmark$  &65.31&76.97&58.43&66.32\\
    TCL && $\checkmark$  & 66.03& 78.14& 59.94& 68.23\\
    \hline
    CLIP & $\checkmark$ & $\checkmark$  & 66.08 & 72.88 & 57.42 & 65.65\\
    DeCLIP & $\checkmark$ & $\checkmark$ &67.16&72.96&58.64&66.49\\
    ALBEF & $\checkmark$ & $\checkmark$ &68.03&74.26&60.04&68.13\\
    TCL & $\checkmark$ & $\checkmark$  & 68.69& 75.02& 60.89& 69.17\\
     \end{tabular}
      }
      \label{tab:tab6}
    \vspace{-1mm}  
\end{table*}

\noindent\textbf{Baselines.}
We consider four baselines: TransE~\cite{TransE}, Q2B~\cite{Q2B} and BETAE~\cite{BetaE&KGreasoning}. Since the \method based plug-in module can be attached to any reasoning models, we implement the Q2B$^*$ with our module based on Q2B and BETAE$^*$ based on BETAE. As shown in Tab.~\ref{tab:4}, BETAE$^*$ achieves on average \textbf{21.64\%} and \textbf{21.23\%} MRR on the validation and testing set of our dataset, respectively. For a fair comparison (\textit{e.g.}, TransE), our dataset does not construct complex logic such as FOL~\cite{computegraph&multihop} to evaluate the performance of multi-hop logical reasoning.

\subsection{Multimodal Event Classification}
\label{sec:uknowtask}
We propose a novel task called multimodal event classification, leveraging event annotations (Tab.~\ref{tab:coraevent}) from both Wiki's event categories and our own manual tagging.
The event annotation helps intelligent machines understand human activities and history, offering the possibility to identify which \textit{type of event} or which \textit{real historical event} a picture or a text is relevant to.
As shown in Tab.~\ref{tab:tab6}, TCL~\cite{TCL} achieves on \textbf{66.80\%} and \textbf{55.87\%} on ACC@1 when using the image-input on the \textit{Event-11} and \textit{Event-9185}, respectively. We simply modify all the baseline methods and add a late-fusion module after the image/text encoder to support multimodal classification. Results show that TCL with multimodal inputs obtains gains of \textbf{1.89\%} and \textbf{5.02\%} compared with the singlemodal, which demonstrates that multimodal pre-training is more helpful for downstream multimodal tasks.

\subsection{Single- \& Cross-Modal Retrieval}
We design four kinds of single- \& cross-modal retrieval tasks: image-to-image, text-to-text, image-to-text, and text-to-image. The construction of GT is based on the event annotations in $G_m$ (Fig.~\ref{fig:event}). We treat images or texts belonging to the same news event as a similar semantic cluster, and the goal of retrieval is to recall the nearest neighbors within this cluster. 
The features used for retrieval are derived from the output of the previous layer of the classifier.

As shown in Tab.~\ref{tab:tab5}, TCL~\cite{TCL} achieves on \textbf{33.24\%}, \textbf{43.37\%} and \textbf{45.22\%} R@1, R@5, R@10 on the zero-shot setting of image retrieval. The results are \textbf{58.89\%}, \textbf{68.47\%} and \textbf{73.91\%} when fine-tuning the pre-trained parameters, which means the pre-training$\rightarrow$fine-tuning strategy is extremely beneficial for downstream retrieval. We provide more details about hyperparameters in Sec.~\ref{sec:Hyperparameters}.

\begin{table*}[ht]
    \tablestyle{5pt}{1.0}
    \def\w{20pt} 
    \setlength\tabcolsep{2pt}
    \caption{\textbf{A new benchmark of the retrieval task.} Zero-shot means freezing the pre-trained parameters then transfer to the test set for inference. Fine-tune means tuning the pre-trained parameters in the training set before inference.
    \vspace{-4mm}
    }      
    \scalebox{1.0}{
      \begin{tabular}{lcccc|ccc}
        \multirow{2}{*}{\bf Model} & \multirow{2}{*}{\bf Retrieval} & \multicolumn{3}{c}{\bf Zero-Shot} & \multicolumn{3}{c}{\bf Fine-Tune}\\
        & & \bf R@1 & \bf R@5 & \bf R@10 & \bf R@1 & \bf R@5 & \bf R@10\\        
    \shline
    CLIP~\cite{CLIP} & IMAGE &  32.41 & 41.96 & 43.92   & 55.97 & 67.44 & 71.28\\
    DeCLIP~\cite{Declip} & IMAGE &  32.75 & 42.36 & 44.38 & 56.96 & 66.59 & 70.95\\
    ALBEF~\cite{ALBEF} & IMAGE &  32.88 & 42.76 & 44.79  & 58.56 & 67.83 & 72.24\\
    TCL~\cite{TCL} & IMAGE & 33.24&43.37&45.22&58.89&68.47&73.91\\
    \hline
    CLIP & TEXT &  33.02 & 42.56 & 46.03  & 56.50 & 65.12 & 70.20\\
    DeCLIP & TEXT & 34.00 & 43.97 & 47.11 & 55.87 & 65.20 & 70.35\\
    ALBEF & TEXT &  33.87 & 43.86 & 46.82 &56.77 & 65.91 & 71.15\\
    TCL & TEXT &  34.67&44.25&47.67&56.60&65.50&70.54\\
    \hline
    CLIP & IMG-to-TXT &  32.73 & 42.64 & 44.72  & 56.32 & 66.93 & 70.61\\
    DeCLIP & IMG-to-TXT & 32.96 & 42.84 & 45.17 & 57.21 & 66.80 & 71.26\\
    ALBEF & IMG-to-TXT &  33.20 & 42.97 & 45.32 & 58.43 & 67.59 & 71.95\\
    TCL & IMG-to-TXT &  33.37&43.25&46.04&58.70&67.88&72.33\\
    \hline
    CLIP & TXT-to-IMG &  31.78 & 41.04 & 42.51  & 55.74 & 64.38 & 69.56\\
    DeCLIP & TXT-to-IMG & 32.13 & 41.55 & 42.99 & 55.84 & 65.12 & 70.32\\
    ALBEF & TXT-to-IMG &  31.95 & 41.32 & 42.85 & 57.21 & 66.04 & 71.50\\
    TCL & TXT-to-IMG &  32.56&42.04&43.74&57.17&65.92&71.47\\
     \end{tabular}
      }
      \label{tab:tab5}
\end{table*}

\subsection{Visual Task Adaptation}
\label{sec:pdtask2}
Visual Task Adaptation Benchmark (VTAB)~\cite{VTAB} is a diverse, realistic, and challenging vision representation benchmark, containing 19 tasks and covering a broad spectrum of domains and semantics. 
These tasks are grouped into three sets: NATURAL, SPECIALIZED, and STRUCTURED which utilize natural world, professional technology and artificial environment images respectively.
We benchmark models on VTAB with ACC@1. We fine-tune models for 10 epoch in each task and compute the inner product between outputs of images and label texts with prompts~\cite{CLIP} through pre-trained image encoders and text encoders as the similarity score.
As shown in Tab.~\ref{tab:tab7}, our approach obtains gains of avg. \textbf{1.14\%} compared with the origin CLIP when fairly using the same \textit{UKnow}'s data for the upstream pre-training.

The backbone of CLIP is ViT-B/32. The cost of pre-train is 26h / 30epoch. The key hyperparameters are \textit{bs: 512, lr: 0.001, warmup: 1e4, eps: 1e-8, beta1: 0.9, beta2: 0.999, dim: 512, AdamW}. The detailed setting can be found in Sec.~\ref{sec:Hyperparameters}.
It is essential to highlight that the image-text PAIR constitutes only one type of data in our protocol. By leveraging the capabilities of \method, our pre-trained CLIP model can effectively comprehend the inherent knowledge ingrained within the data, resulting in superior performance than the original CLIP model (as observed in Tab.~\ref{tab:tab7}, Row2, utilizing image-text PAIR only).

\begin{table*}[!t]
\tablestyle{5pt}{1.0}
\def\w{20pt} 
\setlength\tabcolsep{2pt}
\caption{\textbf{The comparison of \textit{w/} and \textit{w/o} \method pre-training.} Zero means the model is initialized with all-zero parameters \textit{w/o} pre-training. CLIP$^*$ means pre-training with origin CLIP contrast loss on our dataset. Ours means \method pre-training.}
\resizebox{\linewidth}{!}{
  \begin{tabular}{c|ccccccccccccccccccc|c}
    & \rotatebox{90}{CIFAR100} & \rotatebox{90}{Caltech101} & \rotatebox{90}{DTD} & \rotatebox{90}{Flowers102} & \rotatebox{90}{Pets} & \rotatebox{90}{SVHN} & \rotatebox{90}{Sun397} & \rotatebox{90}{Camelyon} & \rotatebox{90}{EuroSAT} & \rotatebox{90}{Resisc45} & \rotatebox{90}{Retinopathy} & \rotatebox{90}{ClevrCount} & \rotatebox{90}{ClevrDist} & \rotatebox{90}{DMLab} & \rotatebox{90}{KITTIDist} & \rotatebox{90}{dSprLoc} & \rotatebox{90}{dSprOri} & \rotatebox{90}{sNORBAzim} & \rotatebox{90}{NORBElev} & \bf \rotatebox{90}{VTAB (avg.)}\\
    \shline
    Zero& 58.39&53.54&49.26&52.51&58.93&64.24&48.96&52.44&63.95&60.03&58.62&62.78&62.59&44.27&45.87&75.89&74.48&67.54&60.89&58.69\\
    CLIP$^*$&75.25&71.74&58.39&77.54&74.40&79.42&61.72&70.42&81.56&76.43&67.85&81.25&80.48&60.03&63.98&84.33&82.66&83.68&76.57&74.09\\
    \rowcolor[gray]{0.95} Ours&76.79&72.73&60.44&78.48&76.33&80.56&62.37&72.23&83.27&77.26&65.91&82.46&81.34&63.37&65.74&85.61&82.79&85.12&76.64&\bf 75.23 \\
  \end{tabular}
  }
  \label{tab:tab7}
    \vspace{-2mm}  
\end{table*}

\subsection{Hyperparameters}
\label{sec:Hyperparameters}

Tab.~\ref{tab:hyperparam1} and Tab.~\ref{tab:hyperparam2} list the hyperparameters that differ on each models and are determined with the validation performance on our dataset. In particular, Tab.~\ref{tab:hyperparam1} lists 7 common hyperparameters, such as learning rate, batch size, warmup, epoch number, \textit{etc.}, employed during pre-training. The pre-trained model is evaluated using a standard pipeline consisting of pre-training on Dataset1, fine-tuning on Dataset2-Train, and testing on either Dataset2-Test/Val. Therefore, we list the hyperparameters used during fine-tuning in Tab.~\ref{tab:hyperparam2}, which are slightly different from Tab.~\ref{tab:hyperparam1}. We omit some of the model results, since ALBEF and TCL share the same set of hyperparameters, and the original CLIP and CLIP-UKnow share the same set of parameters.

\begin{table}[ht]

    \begin{minipage}[b]{1\textwidth}
\tablestyle{5pt}{1.0}
\setlength\tabcolsep{3pt}
\def\w{20pt} 
  \caption{\textbf{Hyperparameters for models of pre-training.}
  } 
\scalebox{1}{
    \begin{tabular}{l|c|c|c}
    \bf Hyperparameter & \bf ALBEF & \bf DeCLIP & \bf CLIP-UKnow \\
    \shline
    \\[-2.5ex]
    Learning Rate    & $0.0001$  & $0.001$   & $0.001$ \\
    Batch Size       & $128$     & $128$     & $512$ \\
    Number of Epochs & $30$      & $30$      & $30$ \\
    Weight Decay     & $0.02$    & $0.1$       & $0.1$ \\
    Optimizer        & $\mathtt{AdamW}$ & $\mathtt{AdamW}$ & $\mathtt{AdamW}$ \\
    Feature Dim      & $256$ & $512$ & $512$ \\
    Warmup           & $20$epc & $5000$ & $10000$ \\  
    \end{tabular}%
      }
  \label{tab:hyperparam1}%
    \end{minipage}
    \hspace{0.8em}
    \begin{minipage}[b]{1\textwidth}
\tablestyle{5pt}{1.0}
\setlength\tabcolsep{3pt}
\def\w{20pt} 
  \caption{\textbf{Hyperparameters for models of fine-tuning.}
  } 
\scalebox{1}{
    \begin{tabular}{l|c|c|c}
    \bf Hyperparameter & \bf ALBEF & \bf DeCLIP & \bf CLIP-UKnow \\
    \shline
    \\[-2.5ex]
    Learning Rate    & $0.0001$  & $5e\text{-}5$   & $5e\text{-}5$ \\
    Batch Size       & $128$     & $256$     & $256$ \\
    Number of Epochs & $128$      & $20$      & $20$ \\
    Weight Decay     & $0.02$    & $0.02$       & $0.02$ \\
    Optimizer        & $\mathtt{AdamW}$ & $\mathtt{AdamW}$ & $\mathtt{AdamW}$ \\
    Feature Dim      & $256$ & $512$ & $512$ \\
    Warmup           & $4$epc & $6$epc & $6$epc \\
    \end{tabular}%
  }
  \label{tab:hyperparam2}%
    \end{minipage}
\end{table}

\subsection{Computation Complexity}
\label{sec:gpu}

Here we detail the time cost of pre-training and fine-tuning. The GPU is NVIDIA(R) A100, the memory of GPU is 81,251MiB, driver version is 470.154, CUDA version is 11.4. The CPU is Intel(R) Xeon(R) Platinum 8369B @ 2.90GHz with 15 physical computation cores. The environment is Python 3.6.12 with Torch 1.10.1. Results are as shown in Tab.~\ref{tab:cc1} and Tab.~\ref{tab:cc2}.
\vspace{-2mm}

\begin{table}[ht]

    \begin{minipage}[b]{1\textwidth}
    \tablestyle{5pt}{0.9}
    \def\w{20pt} 
    \setlength\tabcolsep{1pt}
    \caption{\textbf{The time cost of pre-training.}
    }     
    \scalebox{1.0}{
      \begin{tabular}{lcccc}
       \bf Model & \bf Backbone & \bf Epoch & \bf Batch & \bf Time/h \\
    \shline
    \\[-2ex]
    DeCLIP      & ViT-B/32 & 30 & 128 & 91\\
    ALBEF       & ViT-B/16 & 30 & 128 & 69\\
    TCL         & ViT-B/16 & 30 & 128 & 67\\
    CLIP$^*$    & ViT-B/32 & 30 & 512 & 25\\    
    CLIP-UKnow  & ViT-B/32 & 30 & 512 & 26\\    
     \end{tabular}
      }
      \vspace{2mm}
      \label{tab:cc1}
    \end{minipage}
    \hspace{0.8em}
    \begin{minipage}[b]{1\textwidth}
    \tablestyle{5pt}{0.9}
    \def\w{20pt} 
    \setlength\tabcolsep{1pt}
    \caption{\textbf{The time cost of downstream fine-tuning.}
    }     
    \scalebox{1.0}{
      \begin{tabular}{lcccc|ccc}
        \bf \multirow{2}{*}{Model} & \bf \multirow{2}{*}{Backbone} & \multicolumn{3}{c|}{ \textit{UKnow} Tasks} & \multicolumn{3}{c}{VTAB}\\
       & &\bf Epoch & \bf Batch & \bf Time/h & \bf Epoch & \bf Batch & \bf Time/h\\
    \shline
    \\[-2ex]
    DeCLIP     & ViT-B/32 & 20 & 128 & 12 & - & - & -\\
    ALBEF      & ViT-B/16 & 20 & 128 & 10 & - & - & -\\
    TCL        & ViT-B/16 & 20 & 128 & 10 & - & - & -\\
    Zero$^*$   & ViT-B/32 & - & - & -     & 15 & 128 & 3\\
    CLIP$^*$   & ViT-B/32 & 20 & 256 & 8 & 15 & 128 & 3\\    
    CLIP-UKnow & ViT-B/32 & 20 & 256 & 8 & 15 & 128 & 3\\        
     \end{tabular}
      }
      \label{tab:cc2}
    \end{minipage}
\end{table}

\section{Discussion}

\subsection{Limitation and Future Work}
\label{sec:limitation}
Despite the strides made, our research bears certain limitations. First of all, our current dataset primarily centers on text and image modalities which serve as fundamental pillars for information storage and representation, but lack other useful modalities. In future work, we aim to diversify modalities by augmenting our dataset with a broader range of modalities (\textit{e.g.,} audio, video, 3D, etc.) to facilitate exploration across various downstream tasks. Second, for each downstream task, we selected several basic yet most suitable methods for our work as our baseline, resulting in slight deviations with current state-of-the-art (SOTA) performance. Our primary objective lies in validating the efficacy of our proposed dataset and protocols, and demonstrating the most straightforward and intuitive approach for utilizing our dataset. Hence, we made certain trade-offs, sacrificing some performance by opting for a more rudimentary approach instead of pursuing the SOTA method to enhance understanding and usage. We anticipate that our simplified demonstration will stimulate the community to delve deeper into the potential enhancements that \method can offer in improving performance.

\subsection{Societal Impact}
\label{sec:social_impact}

As stated in Sec.~\ref{sec:da}, our dataset originates from publicly accessible international news sources via the Wikipedia API. These sources only contain events that are publicly available and do not include any sensitive information. Consequently, we confidently affirm that our research carries no potential negative societal impacts.